\documentclass[lettersize,journal]{IEEEtran}
\usepackage{amsmath,amsfonts}
\usepackage{array}
\usepackage[caption=false,font=normalsize,labelfont=sf,textfont=sf]{subfig}
\usepackage{textcomp}
\usepackage{stfloats}
\usepackage{url}
\usepackage{verbatim}
\usepackage{graphicx}
\usepackage{subcaption}
\usepackage{caption}
\usepackage{float} 
\usepackage{xcolor}
\usepackage{multirow}
\usepackage{multicol}
\usepackage{hyperref}
\usepackage{cite}

\usepackage{amssymb}
\usepackage[noend]{algpseudocode}
\usepackage[ruled,vlined]{algorithm2e}
\usepackage{pifont}
\usepackage{hhline}
\newcommand{\NA}{\textcolor{red}{\ding{56}}}

\usepackage{arydshln}
\usepackage{booktabs,colortbl}

\newcolumntype{g}{>{\columncolor[HTML]{EFEFEF}}c}
\newcommand{\AlgStage}[1]{\noindent$\triangleright$~\textbf{#1}\par}

\newcommand{\rev}[1]{#1}

\begin{document}

\title{Graph-Loc: Robust Graph-Based LiDAR Pose Tracking with Compact Structural Map Priors under Low Observability and Occlusion}

\author{Wentao Zhao, Yihe Niu, Zikun Chen, Rui Li, Yanbo Wang, Tianchen Deng, \\ Jingchuan Wang$^{*},~\textit{Senior Member, IEEE}$ 
\thanks{
Wentao Zhao, Zikun Chen, Yanbo Wang, Tianchen Deng, and Jingchuan Wang are with the School of Automation and Intelligent Sensing, Institute of Medical Robotics, Shanghai Jiao Tong University, Shanghai 200240, China.
Yihe Niu is with the School of Mathematical Sciences Shanghai Jiao Tong University.
Rui Li is with Shenzhen Qingmang Robotics Technology Co., Ltd.
Jingchuan Wang (jchwang@sjtu.edu.cn) is the corresponding authors.}
}

\markboth{Journal of \LaTeX\ Class Files,~Vol.~14, No.~8, August~2021}%
{Shell \MakeLowercase{\textit{et al.}}: A Sample Article Using IEEEtran.cls for IEEE Journals}


\maketitle

\begin{abstract}
Map-based LiDAR pose tracking is essential for long-term autonomous operation, where onboard map priors need to be compact for scalable storage and fast retrieval, while online observations are often partial, repetitive, and heavily occluded. We propose Graph-Loc, a graph-based localization framework that tracks the platform pose against compact structural map priors represented as a lightweight point-line graph. Such priors can be constructed from heterogeneous sources commonly available in practice, including polygon outlines vectorized from occupancy/grid maps and CAD/model/floor-plan layouts.
For each incoming LiDAR scan, Graph-Loc extracts sparse point and line primitives to form an observation graph, retrieves a pose-conditioned visible subgraph via LiDAR ray simulation, and performs scan-to-map association through unbalanced optimal transport with a local graph-context regularizer.
The unbalanced formulation relaxes mass conservation, improving robustness to missing, spurious, and fragmented structures under occlusion.
To enhance stability in low-observability segments, we estimate information anisotropy from the refinement normal matrix and defer updates along weakly constrained directions until sufficient constraints reappear.
Experiments on diverse public datasets, controlled stress tests, and real-world deployments demonstrate accurate and stable tracking with KB-level priors from heterogeneous map sources, including under geometrically degenerate structures, dynamic disturbances, gradual scene changes, and sustained occlusion.

\noindent\textit{Note to Practitioners}---Long-term deployment of LiDAR-based localization often requires storing a prior map onboard with strict memory and retrieval constraints, while the robot operates in environments with repetitive structure, partial visibility, and frequent occlusions. Graph-Loc addresses this setting by using a compact structural map representation (a lightweight point-line graph) that can be built from commonly available sources such as occupancy/grid-map outlines and CAD or floor-plan layouts. In operation, the method retrieves only the locally visible portion of the map for matching, which reduces computation and supports fast updates. This design is particularly useful in structured environments (e.g., buildings, corridors, and industrial sites), where geometric observability may temporarily degrade or be blocked by dynamic obstacles. The system outputs incremental platform pose updates and is intended as a drop-in module for navigation stacks that require stable tracking under limited onboard resources. 
\end{abstract}


\section{Introduction}
Reliable LiDAR localization against an onboard map (map-based pose tracking) is a core capability for intelligent vehicles and mobile robots operating over long time horizons.
In practical deployments, two requirements often conflict: the onboard map must be compact for scalable storage and fast retrieval, yet sufficiently discriminative to support robust data association in the presence of real-world clutter and occlusion~\cite{lu2021real,yin2024survey}.
Dense point-cloud maps preserve rich geometry but quickly become expensive to store and query as the covered area grows~\cite{zhang2014loam,vizzo2023kiss,wang2021f}.
This has motivated structural map representations that encode lightweight geometric elements, including polygonal maps \cite{gao2025erpot,gao2018autonomous}, which represent free space and boundaries as compact polygons, as well as road boundaries, curbs, and layout skeletons; such representations are widely adopted in automation and mapping pipelines~\cite{cheng2014topological}.
However, reliable localization with compact structural maps remains challenging when the environment is repetitive, only partially visible, and frequently occluded by surrounding agents~\cite{wijaya2024high}.

A central challenge is that data association becomes fragile when the scene provides weak geometric constraints.
In man-made infrastructures such as parking garages, campuses, depots, and corridors, large portions of a trajectory may be dominated by low-distinctiveness structures, including long straight boundaries and repeated parallel walls~\cite{ebadi2021dare}.
Such geometrically degenerate configurations reduce observability and make greedy correspondence search brittle: incorrect matches can appear plausible, persist over time, and induce substantial drift.
A common remedy in compact-outline-based localization is to split long or irregular primitives into shorter segments to increase matchability~\cite{gao2025erpot}.
While helpful in some cases, splitting inflates the map representation and increases offline processing complexity, and it still struggles when ambiguity is intrinsic due to repetitive layouts and partial observations.
In this work, we assume a fixed compact map available onboard (from offline mapping or existing layouts) and focus on robust scan-to-map pose tracking, without requiring high-level semantic annotations or online map updating.

We address these issues by strengthening the association through graph-level structural matching, rather than increasing feature density. The onboard map is represented as a lightweight point-line graph, and each incoming scan is converted into an observation graph using sparse structural point and line features extracted from range discontinuities and local geometric consistency. Nodes encode geometric features and proximity edges capture local context, turning scan-to-map localization into a graph matching problem where correspondences should be consistent not only per feature but also in their neighborhood relations. This abstraction also accommodates heterogeneous priors commonly available in practice, ranging from polygon outlines vectorized from occupancy/grid-style maps to model/CAD/floor-plan layout priors when accessible.

To compute robust correspondences, we formulate scan-to-map matching as unbalanced optimal transport over graph nodes with a graph-context regularizer.
Unlike nearest-neighbor pipelines that commit to local decisions early~\cite{zhang2014loam,boniardi2019pose,gao2025erpot}, optimal transport estimates a globally consistent transport plan whose assignments are coupled across all nodes.
The unbalanced formulation relaxes strict mass conservation, making the association tolerant to missing structures, spurious returns, and feature fragmentation caused by occlusions and sensing artifacts.

Even with robust matching, low-observability segments can still yield ill-conditioned pose updates.
We therefore introduce a degeneracy-aware delayed optimization strategy: weakly constrained motion directions are detected from information anisotropy, their updates are temporarily deferred while evidence accumulates, and a full update is released once constraints recover.
This stabilizes tracking in structurally degenerate segments and prevents transient association noise from dominating the estimate.


We evaluate Graph-Loc in complementary settings that stress different aspects of map-based pose tracking.
Experiments on KITTI~\cite{geiger2013vision}, the ERPoT dataset~\cite{gao2025erpot}, and \rev{MulRan~\cite{gskim2020mulran}} assess scalability, compact-prior localization accuracy, and robustness under long-term repeated-route changes.
\rev{We further include the Urban Dynamic Objects LiDAR Dataset (DOALS)~\cite{pfreundschuh2021dynamic} to evaluate robustness under real dynamic pedestrian disturbances.}
A controlled simulation based on CMU-EXPLORATION~\cite{cmuexp} is then used \rev{as a diagnostic study} to analyze degeneracy and occlusion under systematically varied pedestrian densities.
Real-world deployments finally evaluate robustness under sustained interference and long-range tracking in an outdoor parking-lot environment, where structural changes introduce mismatches between online observations and a fixed map.

\rev{The central challenge addressed in this work is to maintain reliable localization with compact structural priors when online LiDAR observations are incomplete, dynamically occluded, or locally weakly constrained.}
The main contributions of this paper are summarized as follows:
\begin{itemize}
    \item We propose a compact structural localization framework that represents heterogeneous prior maps and online LiDAR observations as \rev{point-line structural graphs}, enabling pose tracking with sparse but geometrically meaningful priors.

    \item We formulate feature association as a \rev{context-aware unbalanced optimal transport problem}, where graph-context consistency reduces local ambiguity and mass relaxation prevents occluded, missing, or dynamic observations from being forced into incorrect correspondences.

    \item We introduce a \rev{degeneracy-aware delayed optimization strategy} for weakly observable segments, separating well-constrained correction components from temporarily unreliable ones.

    \item We conduct extensive evaluation on public benchmarks, controlled simulation, and real deployments, \rev{validating competitive accuracy with consistently compact maps under occlusion and gradual structural change}.
\end{itemize}

\section{RELATED WORK}
Our work targets practical compact priors for LiDAR localization and studies how to maintain reliable association and stable estimation when observations are weakly informative. We review related work from three aspects: compact structural priors, data association and robust matching, and degeneracy-aware optimization. 

\subsection{Compact Structural Priors}
Most practical LiDAR localization systems use dense point-cloud priors as onboard maps, including LOAM-style pipelines such as ALOAM~\cite{zhang2014loam} and FLOAM~\cite{wang2021f}, as well as KISS\_MCL~\cite{vizzo2023kiss} and HDL\_LOC~\cite{koide2019portable}. 
To improve scalability, these priors are often reduced through submapping and keyframe selection~\cite{tang2019topological,liu2022error}, or replaced by more compact map abstractions such as voxel/implicit encodings~\cite{shan2020lio} and mesh-based representations~\cite{chen2021range}.

\begin{figure*}
      \centering
	  \includegraphics[width=7in]{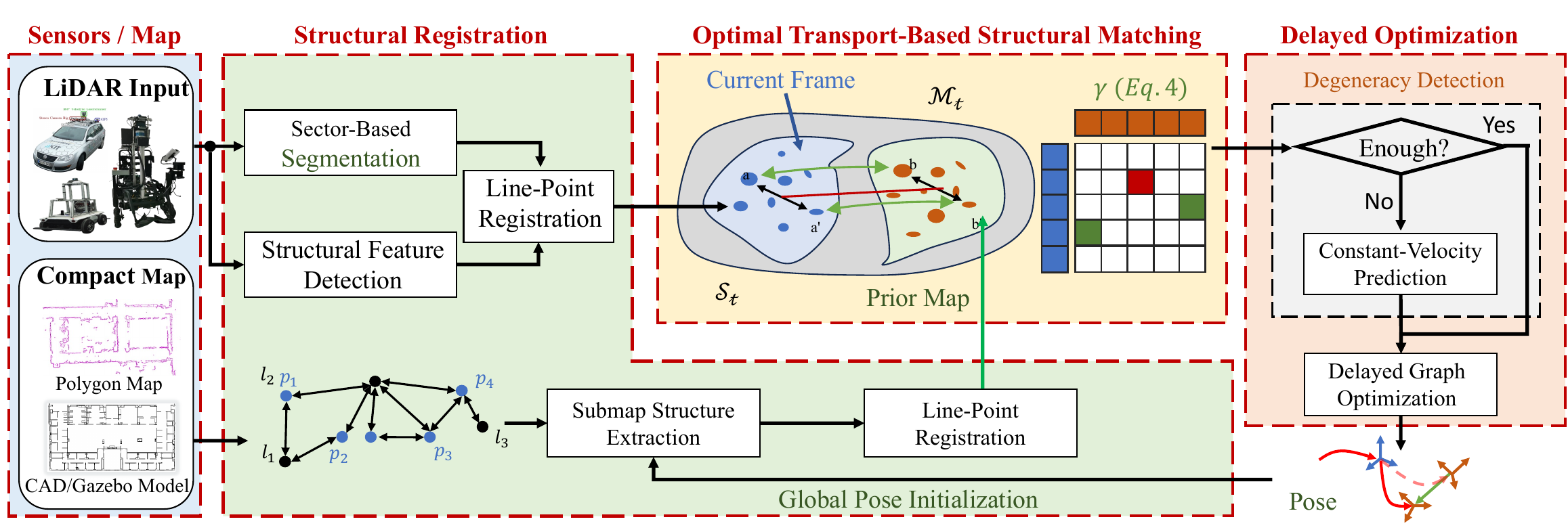}
      \caption{System overview of Graph-Loc. Each LiDAR scan is converted into structural point/line features and organized as an observation graph. A pose-conditioned visible subgraph is retrieved from a compact prior (polygon outlines or CAD/model layouts). Scan-to-map association is solved by unbalanced optimal transport to obtain globally consistent soft correspondences, followed by a degeneracy-aware delayed update for stable real-time pose tracking under occlusion and low observability.}
      \label{Fig_1}
\end{figure*}

Beyond geometric compression, structural priors exploit man-made regularities to produce compact and interpretable abstractions, including CAD/architectural layouts~\cite{boniardi2017robust,boniardi2019pose}, floor-plan assisted localization~\cite{luo2023indoor,xu2022heterogeneous}, and wall/outline extraction via segmentation or layout parsing~\cite{10011847}. Another practical and widely used class is polygonal/outline maps, which derive structure from occupancy/grid-style maps by vectorizing obstacles and boundaries into polylines or polygons. Representative systems such as ERPoT~\cite{gao2025erpot} localize against polygon outlines distilled from offline occupancy maps, achieving strong compactness and fast retrieval. However, compactness alone does not guarantee reliable tracking: in low-distinctiveness or partially observed scenes, long boundaries and repeated structures yield many locally plausible matches, making association fragile. To improve matchability, some outline-based pipelines split long or irregular contours into shorter segments, effectively increasing map density at the cost of additional offline processing and map inflation. In contrast, our work targets the same practicality of compact structural priors but avoids relying on contour splitting by strengthening association through a relational graph abstraction and global matching.

\subsection{Data Association and Robust Matching}
Data association is a key bottleneck for map-based localization, especially when observations are weakly informative due to repetition and occlusions, which frequently occur in real-world deployments~\cite{lu2021real}.
ICP and its variants~\cite{besl1992method,zhang2021fast} and NDT-style methods~\cite{biber2003normal,shafiezadeh2024lidar} are widely used for scan-to-map registration, but they can degrade under partial overlap, clutter, or dynamics, where nearest-neighbor hard correspondences become brittle.
Feature-based pipelines improve efficiency by extracting salient geometric cues; LOAM~\cite{zhang2014loam} is a milestone that uses edge and planar features for registration, and many subsequent systems adopt similar front-ends for odometry and mapping.

Several practical localization systems build global pose tracking on top of such association mechanisms. LOAM-style pipelines (e.g., ALOAM~\cite{zhang2014loam} and FLOAM~\cite{wang2021f}) typically rely on feature-level correspondences and iterative alignment. KISS\_MCL~\cite{vizzo2023kiss} follows an ICP-based registration philosophy designed for robustness with minimal tuning, while HDL\_LOC~\cite{koide2019portable} represents a widely used scan-matching localization stack that combines scan-to-map alignment with filtering. Despite their effectiveness, these pipelines largely depend on hard or locally-coupled correspondence decisions, which can be ambiguous when many matches are locally plausible in repetitive structures or when observations are heavily occluded.

For compact structural priors, however, many pipelines still depend on local correspondence search with hard assignments and heuristic consistency checks.
When the map prior contains repeated or elongated structures and the scan is only partially visible, local matching produces many plausible candidates and can lock onto incorrect correspondences.
This limitation is also reflected in outline-based tracking systems such as ERPoT~\cite{gao2025erpot}, where association is primarily driven by local geometric consistency and can become ambiguous under repetition and partial visibility.
Optimal transport offers an alternative by estimating a global soft assignment between two sets of entities; entropic regularization enables efficient computation~\cite{cuturi2013sinkhorn}, and unbalanced formulations relax mass conservation to tolerate missing and spurious elements under occlusion.
PS\_LOC~\cite{li2024ps} follows this direction by applying optimal-transport-based matching between structural features, improving robustness over greedy local association when outliers and missing features are present.
Building on this line, Graph-Loc formulates scan-to-map association as unbalanced optimal transport on graph nodes with a graph-context regularizer, so correspondences are selected jointly under neighborhood consistency rather than by independent nearest-neighbor decisions, improving robustness without inflating the map prior.

\subsection{Degeneracy Handling and Optimization}
Even with improved association, weakly informative observations can lead to ill-conditioned estimation, where available constraints weakly determine certain motion directions.
Prior work improves stability by incorporating structural regularities such as parallelism and orthogonality~\cite{yunus2021manhattanslam,joo2020linear,li2023hong}, by jointly modeling and associating multiple structural primitives~\cite{li2024ps}, or by using relational/graph abstractions to provide additional context~\cite{gong2021two,shaheer2023graph,zhang2025cornervins}.
Nevertheless, degeneracy can still arise when observations concentrate in a few dominant directions (e.g., long straight boundaries), which is common in corridors, garages, and parking-lot traversals.

This motivates strategies that explicitly detect weak observability and stabilize updates, for example through adaptive regularization or delayed refinement when constraints are insufficient.
Our work follows this direction but targets a more explicit directional treatment: weakly constrained motion directions are detected online and temporarily withheld while evidence accumulates, and a full update is released once observability recovers.
This delayed strategy complements robust optimal-transport-based association and improves reliability under low-informativeness observations.

\section{Method}

\subsection{Problem Statement and Framework}
Figure~\ref{Fig_1} overviews the proposed pipeline for map-based LiDAR pose tracking with a fixed prior map.
We consider ground-vehicle and indoor ground-navigation scenarios (e.g., roads, parking lots, and indoor floors), where the dominant motion is planar.
Accordingly, we perform 3-DoF tracking in the horizontal plane and parameterize the pose by planar translation and yaw:
\begin{equation}
\mathbf{p}_t = [x_t,\ y_t,\ \phi_t]^\top,\qquad \mathbf{P}_t \in \mathrm{SE}(2).
\end{equation}
The LiDAR scan at time $t$ is denoted by $\mathbf{X}_t$.
Our goal is to estimate the global pose $\mathbf{P}_t$ of each scan with respect to the prior map while maintaining reliable scan-to-map association under partial visibility and occlusion.

Starting from a LiDAR front end, each scan is converted into sparse structural measurements.
Specifically, we extract geometric cues from range discontinuities and local surface consistency, and consolidate them into two feature types: point features and line features.
These sparse features provide an efficient abstraction while retaining stable geometric evidence for association.

To incorporate local context without sacrificing compactness, both observations and the prior map are represented as graphs (Sec.~\ref{sec:structural}).
For the current scan, we construct an observation graph $\mathcal{S}_t=(\mathcal{V}_t,\mathcal{E}_t)$, where each node $v\in\mathcal{V}_t$ corresponds to a point or line feature with geometric attributes (e.g., position and orientation).
Edges $\mathcal{E}_t$ encode local neighborhood structure via Euclidean $k$-nearest neighbors based on representative positions (point locations and line anchors).
Likewise, the offline prior map is represented as a fixed graph $\mathcal{M}=(\mathcal{V}^M,\mathcal{E}^M)$.

Given a pose prediction $\hat{\mathbf{P}}_t$ computed from the previous estimate under a constant-velocity motion model (i.e., extrapolating $\mathbf{P}_{t-1}$ by reusing the last incremental motion), we retrieve a pose-conditioned visible subgraph $\mathcal{M}_t\subset\mathcal{M}$ to restrict matching to plausible candidates and reduce distractors.
We then refine the pose by jointly considering (i) graph-level scan-to-map association and (ii) stability under weak geometric constraints:
\begin{equation}
\begin{aligned}
\Delta\boldsymbol{\xi}_t^\star
&=\arg\min_{\Delta\boldsymbol{\xi}\in \mathbb{R}^{3}}
\ \mathcal{L}_{\mathrm{uot}}\!\left(\hat{\mathbf{P}}_{t}\exp(\Delta\boldsymbol{\xi}) \,\middle|\, \mathcal{S}_t,\mathcal{M}_t\right)
+ \mathcal{R}_{\mathrm{deg}}(\Delta\boldsymbol{\xi}),\\
\mathbf{P}_t
&=\hat{\mathbf{P}}_t\exp(\Delta\boldsymbol{\xi}_t^\star).
\end{aligned}
\label{eq:absolute_pose_inc}
\end{equation}
Here $\mathcal{L}_{\mathrm{uot}}$ performs scan-to-map matching via unbalanced optimal transport on graph nodes (Sec.~\ref{sec:matching}), while $\mathcal{R}_{\mathrm{deg}}$ stabilizes refinement in low-observability configurations through degeneracy-aware masking and delayed updates (Sec.~\ref{sec:degeneracy}).
This formulation produces a robust global pose estimate for each scan with respect to the fixed prior map, without requiring explicit chaining of relative motions.

\begin{figure}
    \centering
    \includegraphics[width=\linewidth]{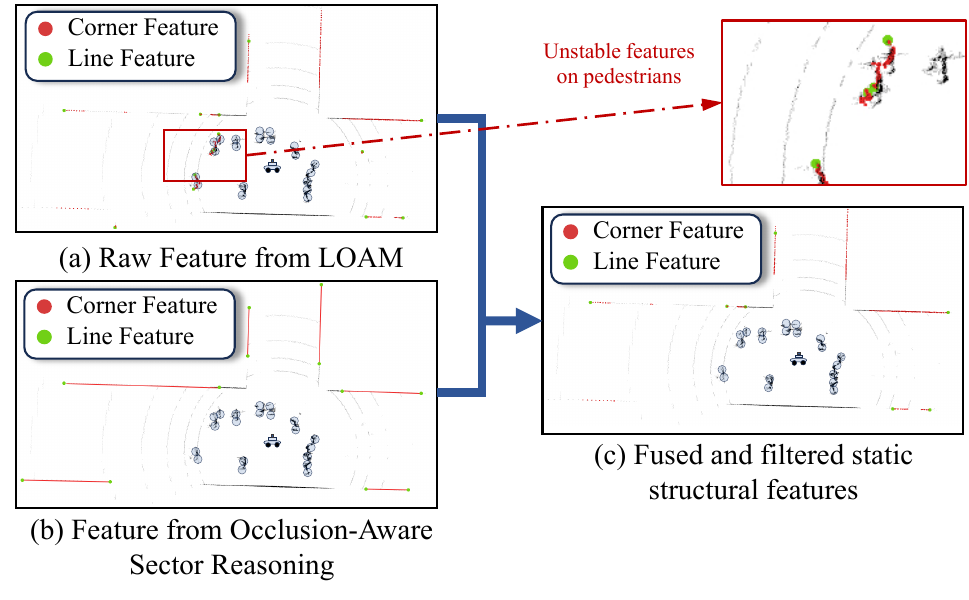}
    \caption{Qualitative results of feature extraction and fusion on high-occlusion environments. 
    (a) LOAM-style short features. (b) Sector-based structural long features. (c) Fused and filtered features used to build $\mathcal{S}_t$.}
    \label{Fig_2}
\end{figure}

\begin{figure*}
      \centering
	  \includegraphics[width=7in]{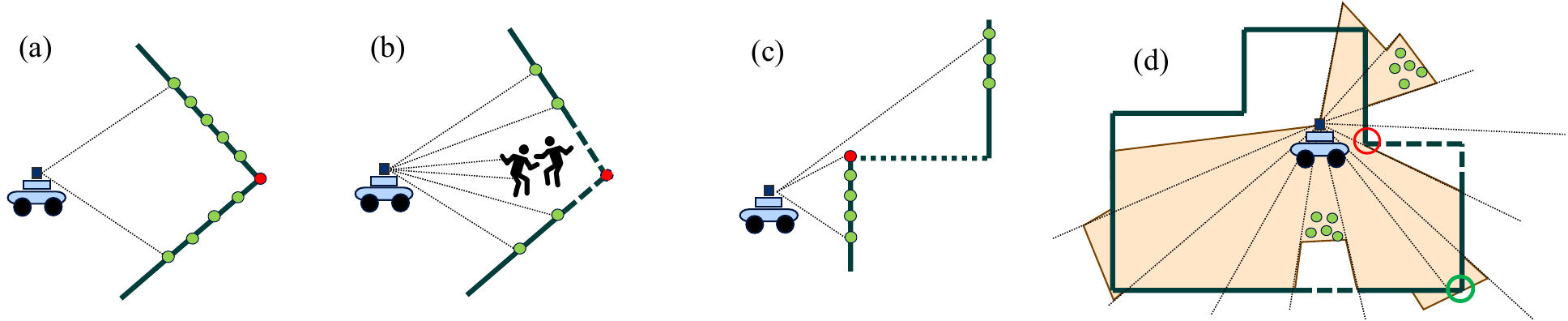}
      \caption{Illustration of structural corner inference under different occlusion scenarios.
      (a) Corners/intersections are not directly observed. (b) Temporary occlusion by dynamic objects.
      (c) Viewpoint-induced structural missingness. (d) Sector-based reasoning parallelized across angular bins.}
      \label{Fig_3}
\end{figure*}

\subsection{Structural Registration}
\label{sec:structural}

Given a LiDAR scan $\mathbf{X}_t$, our goal is to extract a compact yet reliable set of structural features for scan-to-map tracking under partial visibility and dynamic occlusions.
As shown in Figure~\ref{Fig_2}, we build the observation graph via a structural registration module $\mathcal{S}_t \leftarrow \mathrm{Reg}(\mathbf{X}_t)$ that combines two complementary streams: (i) short LOAM-style sparse features and (ii) long sector-based structural lines. We fuse these streams to obtain the final point and line features used for graph construction, with improved robustness to partial visibility and dynamic occlusions.

\subsubsection{Short sparse features from LOAM front-ends}
Following common LiDAR front-ends~\cite{zhang2014loam,gao2025erpot}, we first extract a sparse set of edge-like point features and locally planar/surface cues from $\mathbf{X}_t$.
These features are fast to compute and typically appear in large numbers, providing sufficient local constraints for iterative pose refinement.
However, because they are selected from local neighborhoods, they are sensitive to transient returns caused by close-range occluders and dynamic agents (e.g., points on pedestrians or vehicles), which can dominate the curvature/smoothness statistics and reduce repeatability across frames.
In low-distinctiveness segments (e.g., long straight structures) or under close-range occluders, the neighborhoods used for feature selection can be dominated by transient returns, further reducing robustness.

\subsubsection{Long structural lines from sector-based extraction}
To obtain more repeatable structural cues beyond local neighborhoods, we additionally extract long line features using a sector-based module.
The scan is partitioned into angular sectors, and candidate boundary supports are detected within each sector using range discontinuities and local geometric consistency.
Candidates that are consistent across neighboring sectors are aggregated into line features.
Each line feature $\ell$ is parameterized by two endpoints and a unit direction, and represented in the graph by an anchor point (e.g., midpoint) and its direction.
Compared to LOAM-style short features, long lines are typically more stable across time and viewpoint changes.
However, under occlusion and limited visibility, the observed support is often truncated, which makes endpoints ambiguous and can lead to fragmented or drifting line segments---a key issue that we explicitly address in this work.
These long lines may also provide weaker instantaneous constraints in degenerate layouts (e.g., many near-parallel lines).

A practical difficulty of long lines is that their utility hinges on reliable endpoints.
When a boundary is partially occluded, the visible support becomes truncated and endpoints can drift with the fragment, which may cause erroneous merging of unrelated structures.
We explicitly handle common occlusion patterns (illustrated in Figure~\ref{Fig_3}) during sector aggregation:
(a) Hidden junction cues: when corners/intersections are not directly observed, we infer a soft intersection hypothesis from stable neighboring directions and treat it as weak evidence to stabilize endpoint placement, without making hard commitments.
(b) Truncation under temporary occlusion: we estimate line direction from visible inliers and update endpoints conservatively within verified visible sectors, avoiding aggressive extension beyond observed support.
(c) Viewpoint-induced missing orthogonal walls: in corridor-like scenes, the scan may observe two long parallel side walls, while the orthogonal end wall is outside the field of view, yielding anisotropic constraints. Under a Manhattan-world assumption, we hypothesize a weak orthogonal line to complement the missing structure and use it as soft evidence for stabilizing estimation, without enforcing hard completion when the orthogonal surface is not directly observed.
These strategies improve the repeatability of long lines under occlusion while preventing over-extension when structural endpoints are not observable.

\subsubsection{Fusion for dynamic filtering}
Dynamic objects often produce short-lived returns that break geometric continuity.
To suppress such interference, we fuse the two feature streams:
LOAM-style point features are validated against nearby stable line supports, retaining points that are geometrically consistent with extracted lines (and down-weighting/discarding unsupported ones);
conversely, line candidates without sufficient local support are removed to avoid spurious completion.
This fusion yields a filtered set of point and line features that is more robust to dynamics and partial visibility, and forms the node set of $\mathcal{S}_t$.

\subsubsection{Pose-conditioned prior retrieval via LiDAR ray simulation}
Given the previous estimate $\mathbf{P}_{t-1}$, we form a pose prediction under a constant-velocity assumption by reusing the last incremental motion:
\begin{equation}
\hat{\mathbf{P}}_t=\mathbf{P}_{t-1}\exp(\hat{\boldsymbol{\xi}}_{t-1}),
\qquad
\hat{\boldsymbol{\xi}}_{t-1}=\log\!\left(\mathbf{P}_{t-2}^{-1}\mathbf{P}_{t-1}\right)\in\mathbb{R}^3,
\label{eq:cv_predict}
\end{equation}
where $\log(\cdot)$ maps an $\mathrm{SE}(2)$ transform to its minimal 3-DoF vector representation (planar translation and yaw).
We use $\hat{\mathbf{P}}_t$ to retrieve a pose-conditioned visible subgraph $\mathcal{M}_t\subset\mathcal{M}$ and initialize the subsequent scan-to-map refinement in Eq.~\eqref{eq:absolute_pose_inc}.
Specifically, as shown in Figure~\ref{Fig_3}(d), we simulate LiDAR emission in the map frame by casting rays that follow the sensor's azimuth-elevation sampling pattern.
For each ray originating at the sensor center, we query its first intersection with prior structural elements under a maximum range constraint.
Only the hit elements (and their local neighborhoods) are kept to form $\mathcal{M}_t$.
This visibility-aware retrieval (i) removes distractors outside the current view, reducing ambiguity in repetitive layouts, and
(ii) significantly shrinks the candidate set, accelerating the subsequent matching while preserving local relational consistency within the visible region.

\subsection{Unbalanced Optimal Transport for Feature Matching}
\label{sec:matching}

Given the observation graph $\mathcal{S}_t$ and the retrieved prior subgraph $\mathcal{M}_t$,
correspondences are computed by a single unbalanced optimal transport (UOT) problem on graph nodes.
\rev{The matching objective is constructed progressively from local geometric compatibility to structure-aware and occlusion-tolerant association.}
Pairwise geometric costs first score whether individual candidate feature pairs are locally plausible under the current pose estimate.
\rev{Since feature-level compatibility alone can be ambiguous in repetitive structures, such as corridors, parallel boundaries, or long wall segments, graph-context coupling is introduced to favor match sets that preserve local structural relationships.}
\rev{Finally, mass relaxation addresses a different failure mode: under occlusion or dynamic interference, some prior primitives may be invisible and some observed primitives may be spurious, so they should be allowed to remain unmatched rather than being forced into one-to-one correspondences.}

\rev{This primitive-level formulation also differs from contour- or region-based polygon matching strategies. Graph-Loc does not use polygon-interior tests, polygon centers, or region-based point matching. Since the prior map is represented as a compact point-line structural graph, observation point features are evaluated by geometric compatibility, graph-context consistency, and mass-relaxed UOT, regardless of whether they appear inside a closed outline. This avoids introducing special inside-outline rules and keeps the matching objective unified across point and line primitives. Inconsistent or spurious observations can remain unmatched through the unbalanced formulation, avoiding forced incorrect correspondences.}

Formally, let $\mathcal{G}_x=(\mathcal{V}_x,\mathcal{E}_x)$ and $\mathcal{G}_y=(\mathcal{V}_y,\mathcal{E}_y)$ denote the source and target graphs.
Under a candidate planar pose $\mathbf{P}\in\mathrm{SE}(2)$, we estimate a nonnegative transport plan
$\gamma\in\mathbb{R}_+^{|\mathcal{V}_x|\times|\mathcal{V}_y|}$ by
\begin{equation}
\begin{aligned}
\gamma^\star(\mathbf{P})
=\arg\min_{\gamma\ge 0}\;\;
&\underbrace{\langle \gamma, C(\mathbf{P}) \rangle}_{\text{geometric matching}}
+\beta\,\underbrace{\Omega_{\mathrm{rel}}(\gamma;\mathcal{E}_x,\mathcal{E}_y)}_{\text{context coupling}} \\
&+\rho\,\underbrace{D_{\mathrm{uot}}(\gamma;\mu,\nu)}_{\text{mass relaxation}}
+\varepsilon\,\underbrace{\mathcal{H}(\gamma)}_{\text{entropic smoothing}},
\end{aligned}
\label{eq:uot_graph}
\end{equation}
where $\langle \gamma, C(\mathbf{P}) \rangle=\sum_{i,j}\gamma_{ij}C_{ij}(\mathbf{P})$ scores candidate pairs after transforming observation features into the map frame using $\mathbf{P}$;
$\Omega_{\mathrm{rel}}$ encourages locally consistent match sets through graph context and suppresses structurally inconsistent associations;
the KL-based term relaxes mass conservation so missing, occluded, spurious, or fragmented features are not forced into hard correspondences;
and $\mathcal{H}(\gamma)$ provides entropic smoothing for a stable and efficient Sinkhorn-style solver.

\subsubsection{Geometric matching term $\langle\gamma, C(\mathbf{P})\rangle$}
Nodes are of two types: point features and line features.
Given a candidate planar pose $\mathbf{P}\in\mathrm{SE}(2)$, we transform observation features into the map frame and define type-specific pairwise costs.

For point-point pairs, the cost is the Euclidean distance
\begin{equation}
c_{pp}(i,j;\mathbf{P})=\bigl\|\mathbf{p}_i(\mathbf{P})-\mathbf{p}_j\bigr\|_2,
\label{eq:cpp}
\end{equation}
where $\mathbf{p}_i(\mathbf{P})$ denotes the transformed location of the source point feature under $\mathbf{P}$.

For line-line pairs, each line feature is represented by a unit direction $\mathbf{d}\in\mathbb{S}^1$ and an anchor point $\mathbf{q}\in\mathbb{R}^2$ (e.g., midpoint),
and the cost combines orientation discrepancy and anchor displacement:
\begin{equation}
c_{\ell\ell}(i,j;\mathbf{P})=
w_\theta\,\Delta\theta_{ij}^2
+w_\perp\,\|\mathbf{t}_{ij}^{\perp}\|_2
+w_\parallel\,\|\mathbf{t}_{ij}^{\parallel}\|_2,
\label{eq:cll}
\end{equation}
with $\Delta\theta_{ij}=\arccos\!\bigl(|\mathbf{d}_i(\mathbf{P})^\top\mathbf{d}_j|\bigr)$ (invariant to sign flips),
$\boldsymbol{\delta}_{ij}=\mathbf{q}_i(\mathbf{P})-\mathbf{q}_j$,
$\mathbf{t}_{ij}^{\parallel}=(\boldsymbol{\delta}_{ij}^\top \mathbf{d}_j)\mathbf{d}_j$,
and $\mathbf{t}_{ij}^{\perp}=\boldsymbol{\delta}_{ij}-\mathbf{t}_{ij}^{\parallel}$.
Thus $c_{\ell\ell}$ jointly penalizes orientation mismatch, cross-track offset, and along-track shift.

Furthermore, we additionally allow directional cross-type association from observed point features to map line features.
Specifically, for $v_i\in\mathcal{V}_x^{p}$ (observation) and $v_j\in\mathcal{V}_y^{\ell}$ (map), we define
\begin{equation}
c_{p \ell}(i,j)=\mathrm{dist}\!\left(\mathbf{p}_i,\,\ell_j\right),
\label{eq:cpl}
\end{equation}
where $\mathrm{dist}(\mathbf{p},\ell)$ is the Euclidean point-to-line distance to the (infinite) supporting line of the map line feature $\ell_j$ (or to the finite line feature if a segment distance is used).
This asymmetric channel is useful when the observation provides repeatable point evidence (e.g., corner-like returns) while the corresponding line evidence is weak, partially visible, or fragmented; it also avoids introducing ambiguous map-point hypotheses that are not reliably observable from a single scan.

\subsubsection{Context coupling $\Omega_{\mathrm{rel}}$}
To reduce ambiguity in repetitive layouts, we regularize the plan by preserving local relations.
We use a sparse second-order coupling on source edges $(i,i')\!\in\!\mathcal{E}_x$:
\begin{equation}
\Omega_{\mathrm{rel}}(\gamma)=
\sum_{(i,i')\in\mathcal{E}_x}\ \sum_{j,j'} 
\gamma_{ij}\gamma_{i'j'}\,
\psi\!\left((i,i'),(j,j')\right),
\label{eq:rel}
\end{equation}
where $\psi$ enforces rigid consistency:

(i) Point-point (distance consistency).
\begin{equation}
\psi_{pp}\!\left((i,i'),(j,j')\right)=
\Big(\|\mathbf{p}_i-\mathbf{p}_{i'}\|_2-\|\mathbf{p}_j-\mathbf{p}_{j'}\|_2\Big)^2.
\label{eq:psi_pp}
\end{equation}

(ii) Line-line (angle consistency).
\begin{equation}
\begin{aligned}
\psi_{\ell\ell}\!\left((i,i'),(j,j')\right)
&=
\Big(\theta(\mathbf{d}_i,\mathbf{d}_{i'})
-\theta(\mathbf{d}_j,\mathbf{d}_{j'})\Big)^2, \\
\theta(\mathbf{a},\mathbf{b})
&=\arccos\!\bigl(|\mathbf{a}^\top\mathbf{b}|\bigr).
\end{aligned}
\label{eq:psi_ll}
\end{equation}

We apply $\psi_{pp}$ when $(i,i')$ and $(j,j')$ are point pairs, and $\psi_{\ell\ell}$ when they are line pairs
(otherwise $\psi=0$). $\mathcal{E}_x$ and $\mathcal{E}_y$ are built by Euclidean $k$NN over feature representative positions.

\subsubsection{Mass relaxation $D_{\mathrm{uot}}(\gamma;\mu,\nu)$ and entropic smoothing $\mathcal{H}(\gamma)$}
We adopt a KL-penalized unbalanced formulation that relaxes the marginal constraints:
\begin{equation}
D_{\mathrm{uot}}(\gamma;\mu,\nu)=
\mathrm{KL}(\gamma\mathbf{1}\,\|\,\mu)+\mathrm{KL}(\gamma^\top\mathbf{1}\,\|\,\nu),
\label{eq:uot_kl}
\end{equation}
where $\mathrm{KL}(\mathbf{a}\,\|\,\mathbf{b})=\sum_i a_i\log\frac{a_i}{b_i}-a_i+b_i$ is applied element-wise and $\mathbf{1}$ is an all-ones vector of compatible size.
Unless stated otherwise, we use uniform masses $\mu=\frac{m}{|\mathcal{V}_x|}\mathbf{1}$ and $\nu=\frac{m}{|\mathcal{V}_y|}\mathbf{1}$ with a shared total mass $m$, while the penalty weight $\rho$ controls how strongly the plan adheres to these masses.
This relaxation allows partial matching when features are missing, occluded, spurious, or fragmented, which is common under occlusion and limited visibility.
The entropic term is defined as $\mathcal{H}(\gamma)=-\sum_{i,j}\gamma_{ij}(\log\gamma_{ij}-1)$.
Together with $\varepsilon\,\mathcal{H}(\gamma)$, the KL-based marginal relaxation admits a stable unbalanced Sinkhorn solver, making the global assignment practical online.

\subsubsection{Solver and sparsification}
In practice, Eq.~\eqref{eq:uot_graph} is solved using a Sinkhorn-style iterative scaling algorithm with entropic smoothing, where the unbalanced KL penalties lead to relaxed marginal updates.
To keep computation tractable online, we exploit the sparsity induced by pose-conditioned visible-subgraph retrieval and $k$NN graph construction: candidate pairs $(i,j)$ are restricted to a sparse neighborhood (e.g., within a gating radius or top-$K$ nearest candidates under the geometric cost), and the context coupling $\Omega_{\mathrm{rel}}$ is evaluated only on source edges $(i,i')\in\mathcal{E}_x$ with the corresponding sparse candidate pairs $(j,j')$.
This yields an efficient approximate second-order coupling that retains local-context regularization while keeping both memory and runtime bounded in practice.

\subsection{Degeneracy-Aware Delayed Optimization}
\label{sec:degeneracy}

Even with robust scan-to-map matching, pose refinement can become ill-conditioned when the visible structure provides weak constraints.
For ground navigation, a typical case is a long straight segment (e.g., corridors or parking aisles), where most matched line features are (near-)parallel.
Such configurations constrain lateral translation and yaw well, but weakly constrain translation along the dominant direction.
As a result, per-frame refinement may inject association noise into the weakly constrained mode and accumulate drift.
We therefore detect weak directions online and delay their correction until observability recovers.

After obtaining UOT correspondences, we form a local least-squares refinement problem in 3-DoF (planar translation and yaw) and compute the normal equations
\begin{equation}
\mathbf{H}_t=\mathbf{J}_t^\top \mathbf{W}_t \mathbf{J}_t,\qquad
\mathbf{g}_t=\mathbf{J}_t^\top \mathbf{W}_t \mathbf{r}_t,
\label{eq:normal}
\end{equation}
where $\mathbf{r}_t$ stacks residuals induced by matched features, $\mathbf{J}_t$ is the Jacobian w.r.t.\ the instantaneous correction $\delta\boldsymbol{\xi}_t\in\mathbb{R}^3$, $\mathbf{W}_t$ is a robust weight matrix, and $\mathbf{g}_t$ is the corresponding gradient term in the normal equations.
\rev{The delayed-optimization module tracks the correction flow across degeneracy states. The correction associated with the current normal equations is denoted as $\delta\boldsymbol{\xi}_t$. In degenerate segments, the masked correction $\delta\boldsymbol{\xi}_t^m$ is immediately applied along well-constrained directions. When observability recovers, the current and buffered constraints are aggregated to compute a released delayed correction $\delta\boldsymbol{\xi}_t^d$. The final correction used in the pose update is denoted as $\Delta\boldsymbol{\xi}_t$.}

We perform an eigen-decomposition $\mathbf{H}_t=\mathbf{U}_t \mathrm{diag}(\lambda_1,\lambda_2,\lambda_3)\mathbf{U}_t^\top$ with $\lambda_1\ge\lambda_2\ge\lambda_3$ and identify weakly observable modes by
\begin{equation}
\mathcal{D}_t
=\{\,k\in\{1,2,3\}\mid \lambda_k<\tau_\lambda\,\},
\label{eq:weakdir}
\end{equation}
where $\tau_\lambda$ is a degeneracy threshold (a relative criterion such as $\lambda_k/\lambda_1<\tau_\lambda$ can also be used to reduce scale sensitivity).

\subsubsection{Masked update in degenerate segments}
If $\mathcal{D}_t\neq\emptyset$, we freeze the weak eigen-modes and update only along well-constrained directions.
Specifically, we construct a mask in the eigen-basis
\begin{equation}
\mathbf{M}_t
=\mathbf{U}_t\,\mathrm{diag}(m_1,m_2,m_3)\,\mathbf{U}_t^\top,\qquad
m_k=
\begin{cases}
0,& k\in\mathcal{D}_t,\\
1,& \text{otherwise},
\end{cases}
\label{eq:maskmat}
\end{equation}
and apply a damped solve that suppresses corrections along weak modes:
\begin{equation}
\delta\boldsymbol{\xi}_t^m
=
-\bigl(\mathbf{H}_t+\lambda_r(\mathbf{I}-\mathbf{M}_t)\bigr)^{-1}\mathbf{g}_t
\;\triangleq\;
\mathrm{SolveMasked}(\mathbf{H}_t,\mathbf{g}_t;\mathcal{D}_t),
\label{eq:masked_update}
\end{equation}
where $\lambda_r$ is a damping coefficient for weak-mode suppression.
\rev{In degenerate segments, the final correction is set to $\Delta\boldsymbol{\xi}_t\leftarrow\delta\boldsymbol{\xi}_t^m$, so only well-constrained components are applied immediately.}
In addition, we cache instantaneous information for delayed refinement:
\begin{equation}
\mathcal{B}_t \leftarrow \mathcal{B}_{t-1}\cup\{(\mathbf{H}_t,\mathbf{g}_t)\},
\qquad \text{if } \mathcal{D}_t\neq\emptyset.
\label{eq:buffer}
\end{equation}

\begin{algorithm}[t]
\caption{System pipeline}
\label{alg:pipeline}
\KwIn{$\mathbf{X}_t,\ \mathcal{M},\ \mathbf{P}_{t-1},\ \mathbf{P}_{t-2},\ \mathcal{B}_{t-1}$}
\KwOut{$\mathbf{P}_t,\ \mathcal{B}_t$ \ \textnormal{(and optionally }$\mathbf{T}_{t-1\rightarrow t}\textnormal{)}$}

\AlgStage{Motion prediction (constant velocity):}
$\hat{\boldsymbol{\xi}}_{t-1}\leftarrow \log(\mathbf{P}_{t-2}^{-1}\mathbf{P}_{t-1})$\;
$\hat{\mathbf{P}}_t\leftarrow \mathbf{P}_{t-1}\exp(\hat{\boldsymbol{\xi}}_{t-1})$\;

\AlgStage{Graph registration:}
$\mathcal{S}_t\leftarrow\mathrm{Reg}(\mathbf{X}_t)$\;
$\mathcal{M}_t\leftarrow\mathrm{RayCast}(\mathcal{M},\hat{\mathbf{P}}_t)$\;

\AlgStage{Matching (scan$\rightarrow$map):}
$\gamma^\star \leftarrow \arg\min_{\gamma\ge0}\ \text{Eq.~\eqref{eq:uot_graph}}$\;

\AlgStage{Degeneracy-aware delayed update:}
$\mathbf{H}_t\leftarrow \mathbf{J}_t^\top\mathbf{W}_t\mathbf{J}_t$\;
$\mathbf{g}_t\leftarrow \mathbf{J}_t^\top\mathbf{W}_t\mathbf{r}_t$\;
$\mathcal{D}_t\leftarrow \mathrm{WeakDir}(\mathbf{H}_t)$\;

\eIf{$\mathcal{D}_t\neq\emptyset$}{
    $\mathcal{B}_t\leftarrow \mathcal{B}_{t-1}\cup\{(\mathbf{H}_t,\mathbf{g}_t)\}$\;
    $\delta\boldsymbol{\xi}_t^m\leftarrow \mathrm{SolveMasked}(\mathbf{H}_t,\mathbf{g}_t;\mathcal{D}_t)$\;
    $\Delta\boldsymbol{\xi}_t\leftarrow \delta\boldsymbol{\xi}_t^m$\;
}{
    $\bar{\mathbf{H}}_t\leftarrow \mathbf{H}_t+\sum_{(\mathbf{H},\cdot)\in\mathcal{B}_{t-1}}\mathbf{H}$\;
    $\bar{\mathbf{g}}_t\leftarrow \mathbf{g}_t+\sum_{(\cdot,\mathbf{g})\in\mathcal{B}_{t-1}}\mathbf{g}$\;
    $\delta\boldsymbol{\xi}_t^d\leftarrow-\bar{\mathbf{H}}_t^{-1}\bar{\mathbf{g}}_t$\;
    $\Delta\boldsymbol{\xi}_t\leftarrow \delta\boldsymbol{\xi}_t^d$\;
    $\mathcal{B}_t\leftarrow\emptyset$\;
}

\AlgStage{Pose update:}
$\mathbf{P}_t\leftarrow \hat{\mathbf{P}}_t\exp(\Delta\boldsymbol{\xi}_t)$\;
\If{\textnormal{need incremental output}}{
    $\mathbf{T}_{t-1\rightarrow t}\leftarrow \mathbf{P}_{t-1}^{-1}\mathbf{P}_t$\;
}
\end{algorithm}

\subsubsection{Release delayed refinement when observability recovers}
Once the configuration becomes non-degenerate ($\mathcal{D}_t=\emptyset$), we release the delayed refinement by aggregating buffered evidence:
\begin{equation}
\bar{\mathbf{H}}_t = \mathbf{H}_t + \!\!\sum_{(\mathbf{H},\mathbf{g})\in\mathcal{B}_{t-1}}\!\!\mathbf{H},
\qquad
\bar{\mathbf{g}}_t = \mathbf{g}_t + \!\!\sum_{(\mathbf{H},\mathbf{g})\in\mathcal{B}_{t-1}}\!\!\mathbf{g},
\label{eq:agg_normal}
\end{equation}
and compute the released delayed correction
\begin{equation}
\delta\boldsymbol{\xi}_t^d = -\bar{\mathbf{H}}_t^{-1}\bar{\mathbf{g}}_t
\;\triangleq\; \mathrm{Solve}(\bar{\mathbf{H}}_t,\bar{\mathbf{g}}_t),
\qquad
\mathcal{B}_t\leftarrow\emptyset.
\label{eq:release_update}
\end{equation}
\rev{Here, $\delta\boldsymbol{\xi}_t^d$ is computed from both the current and buffered constraints. When the delayed correction is released, the final correction delivered to the pose update is set to $\Delta\boldsymbol{\xi}_t\leftarrow\delta\boldsymbol{\xi}_t^d$.}

\subsubsection{Motion propagation}
During degenerate segments, we propagate the pose using a constant-velocity prior and apply only the masked refinement increment.
Specifically, we predict the pose by reusing the last motion increment:
\begin{equation}
\hat{\mathbf{P}}_t
\leftarrow
\mathbf{P}_{t-1}\exp(\hat{\boldsymbol{\xi}}_{t-1}),
\qquad
\hat{\boldsymbol{\xi}}_{t-1}=\log\!\left(\mathbf{P}_{t-2}^{-1}\mathbf{P}_{t-1}\right),
\label{eq:cv_propagate_pose}
\end{equation}
and update the estimate by
\begin{equation}
\mathbf{P}_t \leftarrow \hat{\mathbf{P}}_t\exp(\Delta\boldsymbol{\xi}_t),
\label{eq:pose_update_deg}
\end{equation}
where $\Delta\boldsymbol{\xi}_t$ denotes the final correction delivered to the pose update, obtained from either the masked correction or the released delayed correction depending on the degeneracy state.
\rev{In degenerate segments, $\Delta\boldsymbol{\xi}_t=\delta\boldsymbol{\xi}_t^m$; when observability recovers, $\Delta\boldsymbol{\xi}_t=\delta\boldsymbol{\xi}_t^d$.}

The system pipeline is summarized in Algorithm~\ref{alg:pipeline}.

\begin{figure*}
    \centering
    \includegraphics[width=\linewidth]{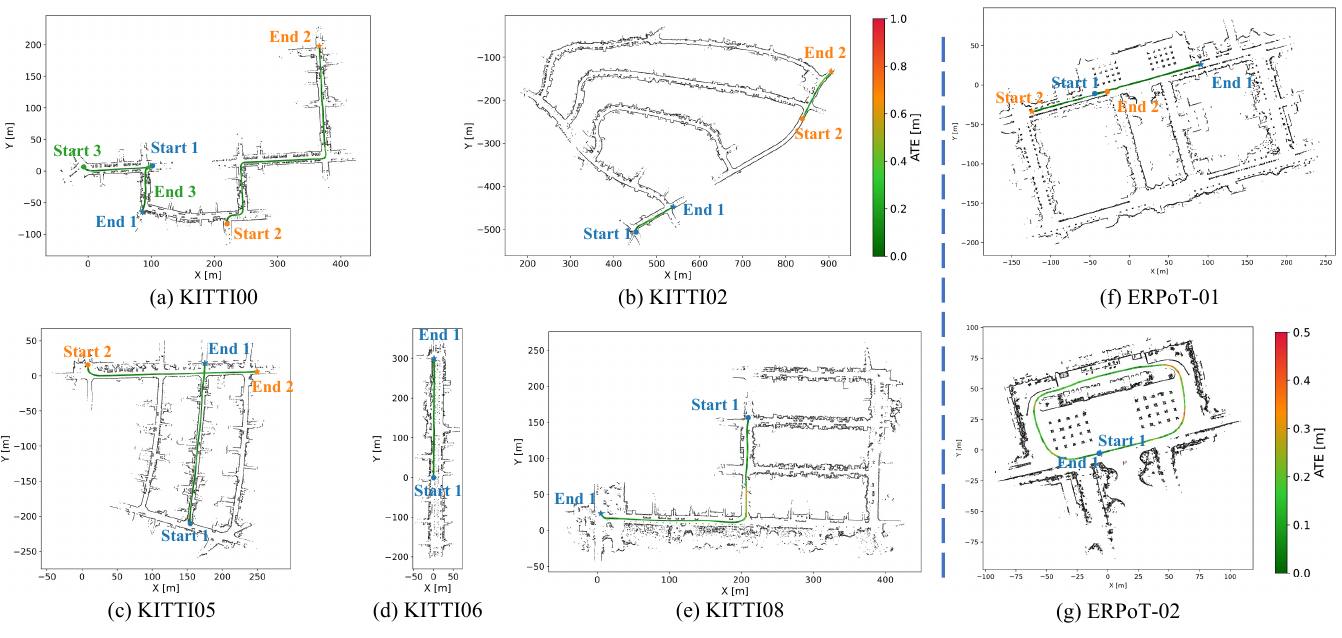}
    \caption{Qualitative results of Graph-Loc on KITTI and ERPoT datasets.}
    \label{Fig_4}
\end{figure*}

\section{EXPERIMENTS}
We evaluate Graph-Loc in complementary settings that stress different aspects of map-based pose tracking.
The experiments are organized into four groups.

First, \textbf{public large-scale benchmarks} evaluate scalability, compact-prior localization accuracy, and robustness under long-term urban changes.
KITTI and ERPoT are used for standard compact-prior localization comparisons, while \rev{MulRan evaluates large-scale long-term repeated-route localization with real LiDAR traversals}.

Second, \textbf{occlusion-robustness experiments} evaluate the method under both real and controlled dynamic disturbances.
\rev{DOALS evaluates real dynamic-object LiDAR sequences with natural pedestrian motion, while CMU-EXPLORATION complements it with a controlled diagnostic setting where pedestrian-induced occlusion density is systematically varied in a degeneracy-prone structural layout.}

Third, \textbf{real-world mobile robot deployments} validate practical tracking under sustained occlusion and gradual scene changes.
The indoor deployment uses CAD-layout and polygon-outline priors to evaluate tracking under repeated straight structures and pedestrian occlusions.
The outdoor parking-lot deployment uses polygon-outline priors to evaluate long-range pose tracking under gradual scene changes, where static priors inevitably deviate from online observations due to parked vehicles, temporary obstacles, and scene reconfiguration.

Finally, \textbf{ablation and runtime studies} analyze the contribution of each module, include a qualitative feature-association comparison, and evaluate the computational efficiency of the system.
Across the above evaluation settings, Graph-Loc is tested with heterogeneous compact priors from different map origins, including Gazebo models, CAD layouts, and occupancy-derived polygon outlines.

\rev{Table~\ref{tab:key_parameters} summarizes the main parameters used in graph construction, UOT-based association, and degeneracy-aware delayed optimization.
Unless otherwise specified, these parameters are fixed across all experiments.}

\begin{table}[t]
\centering
\caption{\rev{Key parameters used in the experiments.}}
\label{tab:key_parameters}
\renewcommand{\arraystretch}{1.15}
\resizebox{\linewidth}{!}{
\begin{tabular}{c c l}
\toprule
Parameter & Value & Meaning \\
\midrule
$k$ & 5 & $k$NN graph edges for local context \\
$K_c$ & 10 & Candidate map nodes per observation node \\
$w_\theta$ & 1.0 & Orientation weight in line-line matching \\
$w_\perp$ & 1.0 & Cross-track offset weight in line-line matching \\
$w_\parallel$ & 0.2 & Along-track offset weight in line-line matching \\
$\beta$ & 0.3 & Graph-context coupling weight in UOT \\
$\rho$ & 0.05 & Mass-relaxation weight in UOT \\
$\varepsilon$ & 0.5 & Entropic smoothing weight in Sinkhorn iterations \\
$\tau_\lambda$ & $10^{-3}$ & Weak-mode threshold for degeneracy detection \\
$\lambda_r$ & 0.1 & Damping coefficient for masked updates \\
\bottomrule
\end{tabular}
}
\end{table}

\begin{table}[t]
\centering
\caption{KITTI splits for offline map construction and pose tracking \cite{gao2025erpot}.}
\label{tab:kitti_splits}
\renewcommand{\arraystretch}{1.15}
\setlength{\tabcolsep}{6pt}
\begin{tabular}{c c c}
\toprule
Seq. & Map construction & Pose tracking \\
\midrule
00 & 0000-1000 & 1:1540-1650, 2:3360-3860, 3:4420-4540 \\
02 & 0920-3408 & 1:4180-4280, 2:4549-4649 \\
05 & 0000-1300 & 1:1301-1581, 2:2300-2650 \\
06 & 0000-0833 & 1:0834-1100 \\
08 & 0000-1306 & 1:1400-1850 \\
\bottomrule
\end{tabular}
\end{table}

\begin{table*}[t]
\caption{Pose-tracking results on the KITTI dataset.}
\label{tab:kitti_results}
\centering
\renewcommand{\arraystretch}{1.25}
\begin{tabular}{ccccccccc}
\hline
Sequence & \multicolumn{2}{c}{Metric} & ALOAM\_MCL & FLOAM\_MCL & KISS\_MCL & HDL\_LOC & ERPoT & Ours \\
\hline

\multirow{4}{*}{KITTI00-1}
& \multicolumn{2}{c}{Prior Map Size (MB)} & 56.44 & 56.44 & 56.44 & 56.44 & \underline{0.28} & \textbf{0.25} \\
\cmidrule(l{4pt}r{4pt}){2-9}
& \multirow{3}{*}{ATE {[}cm{]}}
& Max  & 92.85 & 136.08 & 74.88 & 361.09 & \underline{47.96} & \textbf{22.02} \\
\cmidrule(l{4pt}r{4pt}){3-3}
& & Mean & 58.44 & 61.08 & 49.18 & 24.95 & \underline{13.14} & \textbf{9.36} \\
\cmidrule(l{4pt}r{4pt}){3-3}
& & RMSE & 63.70 & 64.49 & 51.27 & 60.52 & \underline{16.33} & \textbf{10.87} \\
\hline

\multirow{4}{*}{KITTI00-2}
& \multicolumn{2}{c}{Prior Map Size (MB)} & 56.44 & 56.44 & 56.44 & 56.44 & \underline{0.28} & \textbf{0.25} \\
\cmidrule(l{4pt}r{4pt}){2-9}
& \multirow{3}{*}{ATE {[}cm{]}}
& Max  & 158.69 & 138.17 & 208.34 & 149.60 & \textbf{23.66} & \underline{34.01} \\
\cmidrule(l{4pt}r{4pt}){3-3}
& & Mean & 40.92 & 49.73 & 62.20 & \textbf{7.72} & \underline{8.36} & 9.61 \\
\cmidrule(l{4pt}r{4pt}){3-3}
& & RMSE & 47.96 & 58.74 & 81.73 & 11.32 & \textbf{9.53} & \underline{11.27} \\
\hline

\multirow{4}{*}{KITTI00-3}
& \multicolumn{2}{c}{Prior Map Size (MB)} & 56.44 & 56.44 & 56.44 & 56.44 & \underline{0.28} & \textbf{0.25} \\
\cmidrule(l{4pt}r{4pt}){2-9}
& \multirow{3}{*}{ATE {[}cm{]}}
& Max  & 186.50 & 170.32 & 265.02 & 48.52 & \underline{34.46} & \textbf{26.24} \\
\cmidrule(l{4pt}r{4pt}){3-3}
& & Mean & 114.40 & 113.97 & 131.44 & \underline{8.83} & 12.88 & \textbf{7.73} \\
\cmidrule(l{4pt}r{4pt}){3-3}
& & RMSE & 125.01 & 123.72 & 152.89 & \underline{10.79} & 16.64 & \textbf{9.58} \\
\hline

\multirow{4}{*}{KITTI02-1}
& \multicolumn{2}{c}{Prior Map Size (MB)} & 110.47 & 110.47 & 110.47 & 110.47 & \underline{0.85} & \textbf{0.74} \\
\cmidrule(l{4pt}r{4pt}){2-9}
& \multirow{3}{*}{ATE {[}cm{]}}
& Max  & 130.04 & 130.24 & 170.96 & \NA &  \textbf{33.13} & \underline{34.58} \\
\cmidrule(l{4pt}r{4pt}){3-3}
& & Mean & 74.96 & 69.36 & 84.73 & \NA & \textbf{11.58} & \underline{12.22} \\
\cmidrule(l{4pt}r{4pt}){3-3}
& & RMSE & 76.71 & 73.32 & 90.04 & \NA & \textbf{13.39} & \underline{15.49} \\
\hline

\multirow{4}{*}{KITTI02-2}
& \multicolumn{2}{c}{Prior Map Size (MB)} & 110.47 & 110.47 & 110.47 & 110.47 & \underline{0.85} & \textbf{0.74} \\
\cmidrule(l{4pt}r{4pt}){2-9}
& \multirow{3}{*}{ATE {[}cm{]}}
& Max  & \NA & \NA & \NA & 181.80 & \underline{83.44} & \textbf{40.45} \\
\cmidrule(l{4pt}r{4pt}){3-3}
& & Mean & \NA & \NA & \NA & 140.22 & \underline{18.70} & \textbf{14.13} \\
\cmidrule(l{4pt}r{4pt}){3-3}
& & RMSE & \NA & \NA & \NA & 142.50 & \underline{22.10} & \textbf{15.92} \\
\hline

\multirow{4}{*}{KITTI05-1}
& \multicolumn{2}{c}{Prior Map Size (MB)} & 70.93 & 70.93 & 70.93 & 70.93 & \underline{0.36} & \textbf{0.32} \\
\cmidrule(l{4pt}r{4pt}){2-9}
& \multirow{3}{*}{ATE {[}cm{]}}
& Max  & 63.42 & 127.64 & 96.40 & 149.28 & \underline{25.46} & \textbf{24.34} \\
\cmidrule(l{4pt}r{4pt}){3-3}
& & Mean & 32.76 & 40.94 & 48.07 & \underline{8.12} & \textbf{7.94} & 8.94 \\
\cmidrule(l{4pt}r{4pt}){3-3}
& & RMSE & 35.56 & 50.25 & 53.91 & 19.02 & \textbf{9.01} & \underline{10.29} \\
\hline

\multirow{4}{*}{KITTI05-2}
& \multicolumn{2}{c}{Prior Map Size (MB)} & 70.93 & 70.93 & 70.93 & 70.93 & \underline{0.36} & \textbf{0.32} \\
\cmidrule(l{4pt}r{4pt}){2-9}
& \multirow{3}{*}{ATE {[}cm{]}}
& Max  & 101.97 & 216.58 & 117.28 & 103.83 & \underline{26.84} & \textbf{23.07} \\
\cmidrule(l{4pt}r{4pt}){3-3}
& & Mean & 43.56 & 65.23 & 51.24 & 10.68 & \underline{10.14} & \textbf{9.25} \\
\cmidrule(l{4pt}r{4pt}){3-3}
& & RMSE & 50.42 & 85.12 & 60.81 & 13.14 & \underline{11.14} & \textbf{10.39} \\
\hline

\multirow{4}{*}{KITTI06-1}
& \multicolumn{2}{c}{Prior Map Size (MB)} & 56.90 & 56.90 & 56.90 & 56.90 & \underline{0.27} & \textbf{0.24} \\
\cmidrule(l{4pt}r{4pt}){2-9}
& \multirow{3}{*}{ATE {[}cm{]}}
& Max  & \NA & \NA & 547.68 & \NA & \underline{28.78} & \textbf{17.90} \\
\cmidrule(l{4pt}r{4pt}){3-3}
& & Mean & \NA & \NA & 243.60 & \NA & \underline{7.25} & \textbf{6.17} \\
\cmidrule(l{4pt}r{4pt}){3-3}
& & RMSE & \NA & \NA & 276.71 & \NA & \underline{8.08} & \textbf{7.30} \\
\hline

\multirow{4}{*}{KITTI08-1}
& \multicolumn{2}{c}{Prior Map Size (MB)} & 133.01 & 133.01 & 133.01 & 133.01 & \underline{0.59} & \textbf{0.53} \\
\cmidrule(l{4pt}r{4pt}){2-9}
& \multirow{3}{*}{ATE {[}cm{]}}
& Max  & \NA & \NA & \NA & \underline{61.26} & 78.01 & \textbf{46.25} \\
\cmidrule(l{4pt}r{4pt}){3-3}
& & Mean & \NA & \NA & \NA & 24.60 & \textbf{16.10} & \underline{18.71} \\
\cmidrule(l{4pt}r{4pt}){3-3}
& & RMSE & \NA & \NA & \NA & 28.24 & \underline{22.47} & \textbf{21.63} \\
\hhline{=========}
\multirow{3}{*}{Avg.}
& \multirow{3}{*}{ATE {[}cm{]}}
& Max  & \NA & \NA & \NA & \NA & \underline{41.65} & \textbf{27.32} \\
\cmidrule(l{4pt}r{4pt}){3-3}
& & Mean & \NA & \NA & \NA & \NA & \underline{11.90} & \textbf{10.46} \\
\cmidrule(l{4pt}r{4pt}){3-3}
& & RMSE & \NA & \NA & \NA & \NA & \underline{13.81} & \textbf{11.94} \\
\hline

\end{tabular}
\end{table*}

\subsection{Public Large-Scale Benchmarks}
We first evaluate on KITTI, ERPoT, and \rev{MulRan} to assess compact-prior localization across standard benchmark sequences and long-term repeated-route traversals. For each dataset, we follow a prior-map evaluation protocol: an offline prior is built from a dedicated mapping segment or reference traversal, and localization is performed on disjoint tracking segments that overlap in route. All priors are constructed offline and kept fixed during localization.

\subsubsection{KITTI Dataset}
We evaluate on KITTI sequences 00, 02, 05, 06, and 08 following the split protocol in ERPoT~\cite{gao2025erpot};
the exact frame ranges for map construction and pose tracking are summarized in Table~\ref{tab:kitti_splits}.
Table~\ref{tab:kitti_results} highlights the compactness advantage of structural priors:
even after voxel filtering, point-cloud priors typically require tens to hundreds of MB, whereas polygon-outline priors are sub-MB.
Within the polygon setting, our priors are generated from the same outline source but without contour splitting, leading to consistently smaller maps across all evaluated sequences (about 10--13\% reduction) while avoiding additional offline processing.
Despite using these more compact priors, Graph-Loc remains competitive in accuracy.
Compared with all methods, we achieve lower ATE on most tracking segments and obtain the best overall average across all evaluated splits, while keeping the map prior strictly smaller.
Representative trajectories are shown in Figure~\ref{Fig_4}(a)--(e).
Overall, the KITTI results indicate that, in the evaluated compact-prior settings, reliable localization can be achieved without contour splitting when association is solved globally.

\subsubsection{ERPoT Dataset}
We further evaluate on the ERPoT dataset using the official mapping/tracking splits~\cite{gao2025erpot}.
Since the data package for sequence 01-3 is corrupted, we report results on 01-1, 01-2, and 02-1 only.
Table~\ref{tab:erpot_results} summarizes the quantitative results and Figure~\ref{Fig_4}(f)--(g) visualizes representative trajectories.
ERPoT improves matchability by offline contour segmentation that splits long outlines into shorter primitives, whereas we keep the outlines unsplit and instead rely on global unbalanced-OT graph matching together with degeneracy-aware refinement.
Under the same prior source and split protocol, removing contour splitting consistently reduces the map prior size across sequences (about 10--15\%) while preserving robust tracking behavior.

In terms of accuracy, Graph-Loc achieves a lower average ATE than ERPoT across the evaluated sequences.
We observe clear improvements on 01-1 and 01-2.
On 02-1, our error is slightly higher than ERPoT, but the estimated trajectory remains stable and well aligned throughout the sequence, with no divergence or tracking failures.
Overall, these results support that robust global association can reduce reliance on contour splitting: compact polygon-outline priors can remain matchable and achieve competitive, and on average improved, localization accuracy under the standard ERPoT evaluation protocol.

\subsubsection{\rev{MulRan Dataset}}
\rev{We further evaluate on the DCC sequence of the MulRan dataset~\cite{gskim2020mulran} to assess compact-prior localization on large-scale long-term repeated-route LiDAR traversals.}
MulRan provides repeated traversals collected at different times in real urban environments, including temporal changes, lane-level differences, viewpoint variations, and natural traffic participants.
The DCC sequences used in our evaluation span about one month, with Seq.~01, Seq.~02, and Seq.~03 collected on 2019-08-02, 2019-08-23, and 2019-09-03, respectively.
We use DCC Seq.~03 as the reference traversal to construct the fixed compact structural prior, because it provides suitable route coverage for the overlapping portions of the other DCC traversals.
The evaluation follows a prior-map localization protocol rather than a chronological map-update setting: the prior is built offline from the reference traversal and then kept fixed during localization.
We localize on the route-overlap portions of DCC Seq.~01 and DCC Seq.~02.
Because these repeated traversals do not have exactly identical trajectories, we use the first 1800 frames of DCC Seq.~01 and the first 3350 frames of DCC Seq.~02 for localization, avoiding non-overlapping route segments outside the prior-map coverage.
All ATE values are computed on these route-overlap localization segments.
This setting preserves realistic map-query discrepancies caused by temporal changes, viewpoint differences, traffic participants, and route-level variations.

\begin{figure}[t]
    \centering
    \includegraphics[width=\linewidth]{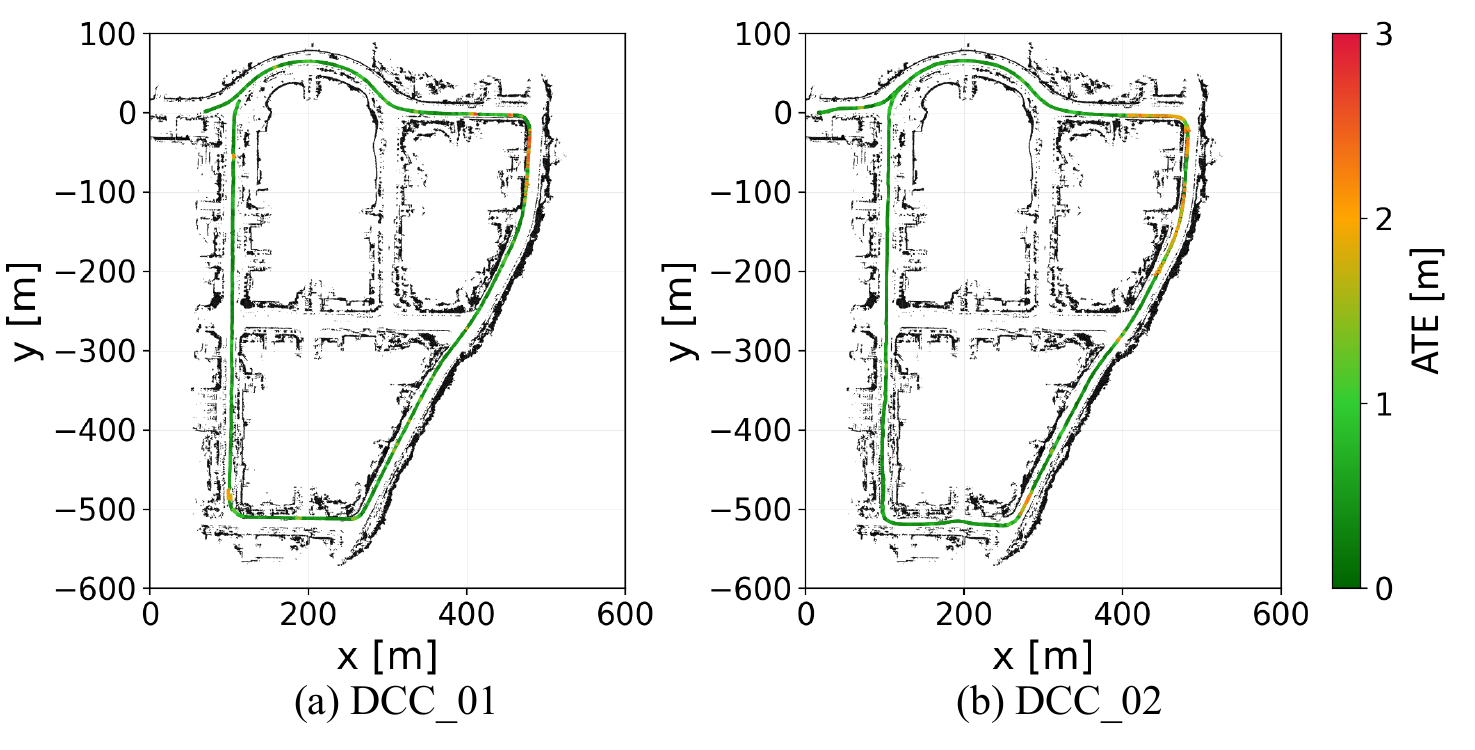}
    \caption{\rev{Representative qualitative results on MulRan DCC.
    The compact prior map is built from DCC Seq.~03, and localization is performed on DCC Seq.~01 and Seq.~02.}}
    \label{Fig_5}
\end{figure}

\rev{As shown in Table~\ref{tab:mulran_results}, Graph-Loc achieves the lowest mean and RMSE errors on both DCC Seq.~01 and Seq.~02, with RMSEs around $61\,\mathrm{cm}$ on both query traversals.
This result suggests that the proposed graph-based association remains effective on long repeated routes with map-query discrepancies, without relying on dense point-cloud priors.
Graph-Loc uses only $0.72\,\mathrm{MB}$ for the compact prior, compared with $3.80\,\mathrm{MB}$ for ERPoT and more than $21\,\mathrm{MB}$ for the point-cloud-prior baselines.
Although ERPoT also uses a compact polygon prior, it fails to provide stable localization in this large-scale setting, indicating that compactness alone is insufficient when scan-to-map association becomes ambiguous.
In contrast, Graph-Loc combines a lightweight prior with robust graph-context UOT association, achieving a better balance between map compactness and localization robustness.
The qualitative trajectories in Figure~\ref{Fig_5} further confirm that Graph-Loc remains stable over the route-overlap portions of the DCC query traversals.}

\begin{table*}[t]
\caption{Pose-tracking results on the ERPoT dataset.}
\label{tab:erpot_results}
\centering
\renewcommand{\arraystretch}{1.25}
\begin{tabular}{ccccccccc}
\hline
Sequence & \multicolumn{2}{c}{Metric} & ALOAM\_MCL & FLOAM\_MCL & KISS\_MCL & HDL\_LOC & ERPoT & Ours \\
\hline

\multirow{4}{*}{ERPoT-01-1}
& \multicolumn{2}{c}{Prior Map Size (MB)} & 111.91 & 111.91 & 111.91 & 111.91 & \underline{0.98} & \textbf{0.87} \\
\cmidrule(l{4pt}r{4pt}){2-9}
& \multirow{3}{*}{ATE [cm]} & Max  & 53.49 & 109.57 & 87.40 & \NA & \textbf{22.26} & \underline{30.42} \\
\cmidrule(l{4pt}r{4pt}){3-3}
&  & Mean & 24.30 & 33.12 & 27.74 & \NA & \underline{9.48} & \textbf{5.39} \\
\cmidrule(l{4pt}r{4pt}){3-3}
&  & RMSE & 26.08 & 40.95 & 32.71 & \NA & \underline{10.24} & \textbf{6.57} \\
\hline

\multirow{4}{*}{ERPoT-01-2}
& \multicolumn{2}{c}{Prior Map Size (MB)} & 111.91 & 111.91 & 111.91 & 111.91 & \underline{0.98} & \textbf{0.87} \\
\cmidrule(l{4pt}r{4pt}){2-9}
& \multirow{3}{*}{ATE [cm]} & Max  & 27.28 & 42.27 & 37.07 & 21.03 & \underline{28.83} & \textbf{16.29} \\
\cmidrule(l{4pt}r{4pt}){3-3}
&  & Mean & 13.14 & 14.44 & 22.11 & 14.05 & \underline{6.97} & \textbf{4.82} \\
\cmidrule(l{4pt}r{4pt}){3-3}
&  & RMSE & 14.03 & 15.35 & 23.59 & 14.33 & \underline{7.99} & \textbf{5.47} \\
\hline

\multirow{4}{*}{ERPoT-02-1}
& \multicolumn{2}{c}{Prior Map Size (MB)} & 32.92 & 32.92 & 32.92 & 32.92 & \underline{0.65} & \textbf{0.56} \\
\cmidrule(l{4pt}r{4pt}){2-9}
& \multirow{3}{*}{ATE [cm]} & Max  & 102.32 & 53.06 & 92.98 & 50.67 & \textbf{29.62} & \underline{32.89} \\
\cmidrule(l{4pt}r{4pt}){3-3}
&  & Mean & 23.09 & 16.47 & 20.29 & \textbf{4.81} & \underline{6.33} & 8.79 \\
\cmidrule(l{4pt}r{4pt}){3-3}
&  & RMSE & 30.57 & 20.15 & 26.28 & \textbf{6.18} & \underline{7.55} & 10.61 \\
\hhline{=========}
\multirow{3}{*}{Avg.}
& \multirow{3}{*}{ATE [cm]}
& Max  & 61.03 & 68.30 & 72.48 & \NA   & \underline{26.90} & \textbf{26.53} \\
\cmidrule(l{4pt}r{4pt}){3-3}
& & Mean & 20.18 & 21.34 & 23.38 & \NA   & \underline{7.59}  & \textbf{6.33} \\
\cmidrule(l{4pt}r{4pt}){3-3}
& & RMSE & 23.56 & 25.48 & 27.53 & \NA   & \underline{8.59}  & \textbf{7.55} \\
\hline

\end{tabular}
\end{table*}

\begin{table}[t]
\caption{\rev{Pose tracking results on MulRan DCC. The prior map is built from DCC Seq.~03, and DCC Seq.~01/02 are used for localization.}}
\label{tab:mulran_results}
\centering
\renewcommand{\arraystretch}{1.25}
\resizebox{\linewidth}{!}{
\begin{tabular}{ccccccc}
\hline
Sequence & \multicolumn{2}{c}{Metric} & HDL\_LOC & ALOAM\_MCL & ERPoT & Ours \\
\hline

\multirow{4}{*}{\begin{tabular}[c]{@{}c@{}}DCC\\Seq.~01\end{tabular}}
& \multicolumn{2}{c}{Prior Map Size (MB)} & 21.43 & 21.39 & \underline{3.80} & \textbf{0.72} \\
\cmidrule(l{4pt}r{4pt}){2-7}
& \multirow{3}{*}{ATE [cm]} & Max  & \textbf{245.32} & 448.72 & \NA & \underline{279.89} \\
\cmidrule(l{4pt}r{4pt}){3-3}
& & Mean & 86.53 & \underline{83.99} & \NA & \textbf{46.35} \\
\cmidrule(l{4pt}r{4pt}){3-3}
& & RMSE & \underline{95.60} & 97.92 & \NA & \textbf{61.65} \\
\hline

\multirow{4}{*}{\begin{tabular}[c]{@{}c@{}}DCC\\Seq.~02\end{tabular}}
& \multicolumn{2}{c}{Prior Map Size (MB)} & 21.43 & 21.39 & \underline{3.80} & \textbf{0.72} \\
\cmidrule(l{4pt}r{4pt}){2-7}
& \multirow{3}{*}{ATE [cm]} & Max  & 309.33 & \underline{297.57} & \NA & \textbf{252.79} \\
\cmidrule(l{4pt}r{4pt}){3-3}
& & Mean & 98.33 & \underline{93.51} & \NA & \textbf{45.20} \\
\cmidrule(l{4pt}r{4pt}){3-3}
& & RMSE & 113.33 & \underline{108.28} & \NA & \textbf{60.55} \\
\hhline{=======}

\multirow{3}{*}{Avg.}
& \multirow{3}{*}{ATE [cm]} & Max  & \underline{277.33} & 373.15 & \NA & \textbf{266.34} \\
\cmidrule(l{4pt}r{4pt}){3-3}
& & Mean & 92.43 & \underline{88.75} & \NA & \textbf{45.78} \\
\cmidrule(l{4pt}r{4pt}){3-3}
& & RMSE & 104.47 & \underline{103.10} & \NA & \textbf{61.10} \\
\hline

\end{tabular}
}
\end{table}

\begin{table}[t]
\caption{Pose tracking results on the Urban Dynamic Objects LiDAR Dataset (DOALS).}
\label{tab:doals_results}
\centering
\renewcommand{\arraystretch}{1.25}
\resizebox{\linewidth}{!}{
\begin{tabular}{ccccccc}
\hline
Sequence & \multicolumn{2}{c}{Metric} & HDL\_LOC & ALOAM\_MCL & ERPoT & Ours \\
\hline

\multirow{4}{*}{\begin{tabular}[c]{@{}c@{}}Hauptgebaeude\\Seq. 01\end{tabular}}
& \multicolumn{2}{c}{Prior Map Size (KB)} & 4743.16 & 3611.58 & \underline{66.3} & \textbf{44.6} \\
\cmidrule(l{4pt}r{4pt}){2-7}
& \multirow{3}{*}{ATE [cm]} & Max  & 94.35 & 25.92 & \underline{25.44} & \textbf{24.34} \\
\cmidrule(l{4pt}r{4pt}){3-3}
& & Mean & \underline{6.41} & 8.43 & 13.27 & \textbf{5.23} \\
\cmidrule(l{4pt}r{4pt}){3-3}
& & RMSE & 9.83 & \underline{9.62} & 14.11 & \textbf{6.31} \\
\hline

\multirow{4}{*}{\begin{tabular}[c]{@{}c@{}}Hauptgebaeude\\Seq. 02\end{tabular}}
& \multicolumn{2}{c}{Prior Map Size (KB)} & 4579.32 & 3462.88 & \underline{67.6} & \textbf{49.3} \\
\cmidrule(l{4pt}r{4pt}){2-7}
& \multirow{3}{*}{ATE [cm]} & Max  & 105.71 & \underline{41.93} & 43.68 & \textbf{34.67} \\
\cmidrule(l{4pt}r{4pt}){3-3}
& & Mean & 6.84 & \textbf{4.94} & 14.93 & \underline{6.31} \\
\cmidrule(l{4pt}r{4pt}){3-3}
& & RMSE & 15.07 & \textbf{6.14} & 16.99 & \underline{8.28} \\
\hline

\multirow{4}{*}{\begin{tabular}[c]{@{}c@{}}Niederdorf\\Seq. 01\end{tabular}}
& \multicolumn{2}{c}{Prior Map Size (KB)} & 5444.22 & 5591.04 & \underline{150.1} & \textbf{107.2} \\
\cmidrule(l{4pt}r{4pt}){2-7}
& \multirow{3}{*}{ATE [cm]} & Max  & 82.58 & 26.38 & \underline{22.28} & \textbf{21.12} \\
\cmidrule(l{4pt}r{4pt}){3-3}
& & Mean & \underline{5.03} & \textbf{4.94} & 6.16 & 5.54 \\
\cmidrule(l{4pt}r{4pt}){3-3}
& & RMSE & 7.45 & \textbf{6.14} & 7.11 & \underline{6.47} \\
\hline

\multirow{4}{*}{\begin{tabular}[c]{@{}c@{}}Niederdorf\\Seq. 02\end{tabular}}
& \multicolumn{2}{c}{Prior Map Size (KB)} & 5413.24 & 5451.77 & \underline{150.0} & \textbf{105.5} \\
\cmidrule(l{4pt}r{4pt}){2-7}
& \multirow{3}{*}{ATE [cm]} & Max  & 185.98 & \underline{24.15} & 25.38 & \textbf{18.30} \\
\cmidrule(l{4pt}r{4pt}){3-3}
& & Mean & 5.98 & \underline{5.65} & 5.99 & \textbf{4.61} \\
\cmidrule(l{4pt}r{4pt}){3-3}
& & RMSE & 12.04 & \underline{6.74} & 6.78 & \textbf{5.44} \\
\hline

\multirow{4}{*}{\begin{tabular}[c]{@{}c@{}}Shopville\\Seq. 01\end{tabular}}
& \multicolumn{2}{c}{Prior Map Size (KB)} & 3184.67 & 4026.36 & \underline{138.7} & \textbf{102.5} \\
\cmidrule(l{4pt}r{4pt}){2-7}
& \multirow{3}{*}{ATE [cm]} & Max  & 91.11 & \textbf{17.24} & 26.89 & \underline{18.43} \\
\cmidrule(l{4pt}r{4pt}){3-3}
& & Mean & 7.44 & \underline{6.98} & 11.32 & \textbf{6.03} \\
\cmidrule(l{4pt}r{4pt}){3-3}
& & RMSE & 10.27 & \underline{7.68} & 12.32 & \textbf{6.63} \\
\hline

\multirow{4}{*}{\begin{tabular}[c]{@{}c@{}}Shopville\\Seq. 02\end{tabular}}
& \multicolumn{2}{c}{Prior Map Size (KB)} & 3151.20 & 3919.87 & \underline{154.0} & \textbf{111.0} \\
\cmidrule(l{4pt}r{4pt}){2-7}
& \multirow{3}{*}{ATE [cm]} & Max  & 100.89 & \textbf{19.98} & 45.04 & \underline{22.48} \\
\cmidrule(l{4pt}r{4pt}){3-3}
& & Mean & \underline{5.40} & \textbf{5.21} & 8.34 & 6.68 \\
\cmidrule(l{4pt}r{4pt}){3-3}
& & RMSE & 8.11 & \textbf{5.80} & 9.65 & \underline{8.05} \\
\hline

\multirow{4}{*}{\begin{tabular}[c]{@{}c@{}}Station\\Seq. 01\end{tabular}}
& \multicolumn{2}{c}{Prior Map Size (KB)} & 8909.28 & 8359.93 & \underline{127.4} & \textbf{87.2} \\
\cmidrule(l{4pt}r{4pt}){2-7}
& \multirow{3}{*}{ATE [cm]} & Max  & \NA & 79.99 & \underline{30.83} & \textbf{22.81} \\
\cmidrule(l{4pt}r{4pt}){3-3}
& & Mean & \NA & \underline{6.75} & 10.69 & \textbf{6.72} \\
\cmidrule(l{4pt}r{4pt}){3-3}
& & RMSE & \NA & \underline{8.86} & 11.70 & \textbf{7.83} \\
\hline

\multirow{4}{*}{\begin{tabular}[c]{@{}c@{}}Station\\Seq. 02\end{tabular}}
& \multicolumn{2}{c}{Prior Map Size (KB)} & 8310.65 & 8204.28 & \underline{123.3} & \textbf{79.9} \\
\cmidrule(l{4pt}r{4pt}){2-7}
& \multirow{3}{*}{ATE [cm]} & Max  & 111.84 & 87.75 & \underline{34.54} & \textbf{16.38} \\
\cmidrule(l{4pt}r{4pt}){3-3}
& & Mean & \textbf{5.53} & \textbf{5.53} & 9.56 & \underline{5.69} \\
\cmidrule(l{4pt}r{4pt}){3-3}
& & RMSE & 8.50 & \underline{8.42} & 10.60 & \textbf{6.64} \\
\hhline{=======}

\multirow{3}{*}{Avg.}
& \multirow{3}{*}{ATE [cm]} & Max  & \NA & 40.42 & \underline{31.76} & \textbf{22.32} \\
\cmidrule(l{4pt}r{4pt}){3-3}
& & Mean & \NA & \underline{6.05} & 10.03 & \textbf{5.85} \\
\cmidrule(l{4pt}r{4pt}){3-3}
& & RMSE & \NA & \underline{7.43} & 11.16 & \textbf{6.96} \\
\hline

\end{tabular}
}
\end{table}

\begin{figure*}[t]
    \centering
    \includegraphics[width=\linewidth]{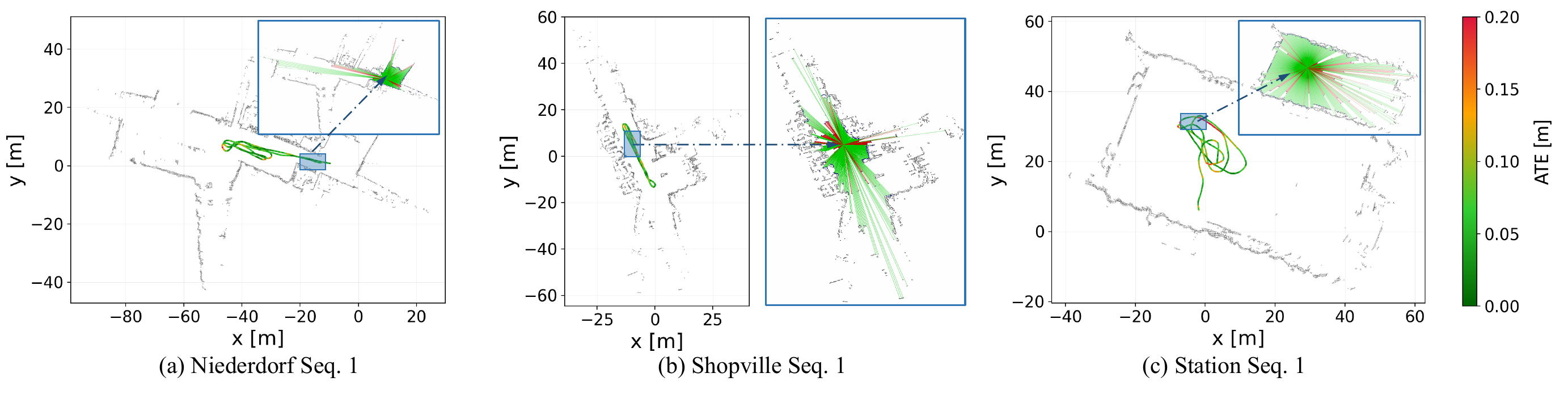}
    \caption{\rev{Representative qualitative results on DOALS.
    The trajectories are overlaid with LiDAR observations and dynamic-object regions in representative Niederdorf, Shopville, and Station sequences.
    The results illustrate pose tracking under real pedestrian-induced occlusions and transient dynamic objects.}}
    \label{Fig_6}
\end{figure*}

\begin{table*}[]
\centering
\caption{Pose-tracking results on CMU-EXPLORATION.
ERPoT* denotes ERPoT with length-based contour splitting for improved matchability in weakly constrained layouts.}
\label{tab:sim_results}
\resizebox{\linewidth}{!}{
\begin{tabular}{c l c *{9}{c}}
\toprule
\multirow{2}{*}{Map Type} &
\multirow{2}{*}{Method} &
\multirow{2}{*}{\begin{tabular}[c]{@{}c@{}}Prior Map\\ Size (KB) $\downarrow$\end{tabular}} &
\multicolumn{3}{c}{Static (ATE [cm])} &
\multicolumn{3}{c}{Mild Occlusion (ATE [cm])} &
\multicolumn{3}{c}{Heavy Occlusion (ATE [cm])} \\
\cmidrule(l{2pt}r{2pt}){4-6}
\cmidrule(l{2pt}r{2pt}){7-9}
\cmidrule(l{2pt}r{2pt}){10-12}
& & &
Max $\downarrow$ & Mean $\downarrow$ & RMSE $\downarrow$
& Max $\downarrow$ & Mean $\downarrow$ & RMSE $\downarrow$
& Max $\downarrow$ & Mean $\downarrow$ & RMSE $\downarrow$ \\
\midrule
\multirow{2}{*}{\begin{tabular}[c]{@{}c@{}}Point\\Cloud\end{tabular}}
& HDL\_LOC    
& 3027.21      
& 97.12 & 8.31 & 9.81
& 97.30 & 8.40 & 9.66
& \NA   & \NA  & \NA \\
& ALOAM\_MCL   
& 3027.21 
& \underline{16.69}  & 9.22             & 9.53
& \underline{16.90}  & 9.26 & \underline{9.57}
& \textbf{26.87}     & \underline{9.90} & \underline{11.05} \\
\noalign{\vskip 2pt}\cdashline{1-12}\noalign{\vskip 2pt}
\multirow{2}{*}{\begin{tabular}[c]{@{}c@{}}Gazebo\\Model\end{tabular}}
& PS\_LOC 
& 4.5 
& 38.24 & 13.32 & 15.05
& 40.98 & 23.27 & 25.12
& \NA   & \NA   & \NA \\
& \cellcolor{gray!15}Ours         
& \cellcolor{gray!15}4.5 
& \cellcolor{gray!15}\textbf{14.78} & \cellcolor{gray!15}\underline{6.40} & \cellcolor{gray!15}\textbf{7.34}
& \cellcolor{gray!15}\textbf{15.88} & \cellcolor{gray!15}\textbf{6.41} & \cellcolor{gray!15}\textbf{7.40}
& \cellcolor{gray!15}\NA            & \cellcolor{gray!15}\NA           & \cellcolor{gray!15}\NA \\
\noalign{\vskip 2pt}\cdashline{1-12}\noalign{\vskip 2pt}
\multirow{3}{*}{\begin{tabular}[c]{@{}c@{}}Polygon\\Map\end{tabular}}
& ERPoT        
& 8.0     
& \NA  & \NA & \NA
& \NA  & \NA & \NA
& \NA  & \NA & \NA \\
& ERPoT*       
& 43     
& 38.34 & 8.04 & 11.98
& \NA   & \NA  & \NA
& \NA   & \NA  & \NA \\
& \cellcolor{gray!15}Ours         
& \cellcolor{gray!15}7.2 
& \cellcolor{gray!15}25.50 & \cellcolor{gray!15}\textbf{6.12} & \cellcolor{gray!15}\underline{8.09}
& \cellcolor{gray!15}29.21 & \cellcolor{gray!15}\underline{7.94} & \cellcolor{gray!15}9.86
& \cellcolor{gray!15}\underline{54.13} & \cellcolor{gray!15}\textbf{9.05} & \cellcolor{gray!15}\textbf{10.64} \\
\bottomrule
\end{tabular}
}
\end{table*}

\subsection{\rev{Robustness to Dynamic Occlusions}}
\rev{This section evaluates occlusion robustness in two complementary settings: DOALS provides real dynamic-object LiDAR sequences, and CMU-EXPLORATION provides controlled simulated sequences for analyzing pedestrian-induced occlusion density in a degeneracy-prone structural layout.}

\subsubsection{\rev{Urban Dynamic Objects LiDAR Dataset (DOALS)}}
\rev{We first evaluate on DOALS~\cite{pfreundschuh2021dynamic}, which contains real LiDAR scans collected in crowded urban and semi-indoor environments with substantial pedestrian motion.}
We use the 10 annotated scans to construct the static compact structural prior and evaluate localization on the remaining scans from the same environment, where pedestrians and other transient objects introduce natural map-observation discrepancies.
Table~\ref{tab:doals_results} reports the quantitative results, and Figure~\ref{Fig_6} shows representative trajectories and dynamic scenes.

\rev{Across the evaluated DOALS sequences, Graph-Loc achieves the best average ATE while using substantially smaller compact priors than point-cloud localization baselines.
DOALS contains real pedestrian motion that introduces transient returns, partial occlusions, and local map-observation inconsistencies.
Compared with ERPoT, our unsplit structural priors remain smaller and achieve lower average error, indicating that the proposed dynamic filtering and unbalanced graph association can maintain reliable compact-prior tracking without forcing transient or occluded observations into incorrect correspondences.
These results validate the robustness of Graph-Loc to real dynamic-object disturbances on real LiDAR data.}

\subsubsection{\rev{Controlled CMU-EXPLORATION Simulation}}
\rev{Complementing the real dynamic-object evaluation on DOALS, we further conduct a controlled study in the CMU-EXPLORATION environment~\cite{cmuexp} to analyze compact-prior pose tracking under systematically varied pedestrian-induced occlusion density in a degeneracy-prone structural layout.} This environment contains long and repeated structures that often yield low-distinctiveness measurements for scan-to-map alignment, and a social-force-based pedestrian simulator enables systematic control of dynamic occlusions.

\textbf{Simulation configuration.}
\rev{To reduce overly idealized simulation assumptions, Gaussian noise with a standard deviation of $0.02 \mathrm{m}$ is added to the simulated LiDAR range measurements.
Dynamic pedestrians are generated by a social-force model with randomized walking speed, body size, and interaction radius, producing goal-directed motion, mutual avoidance, obstacle avoidance, and non-uniform lateral perturbations rather than predefined straight-line trajectories.}
Following the same protocol as the public benchmarks, we build a fixed prior map offline and evaluate localization on three tracking sequences that are disjoint from the mapping run.
The three tracking sequences replay the same route under static, mild-occlusion, and heavy-occlusion settings.
\rev{These settings are generated under the same route, map, and structural layout by progressively reducing the pedestrian population from the heavy-occlusion setting to the mild-occlusion and static settings, corresponding to $20$, $5$, and $0$ simulated pedestrians, respectively.}

\begin{figure}
    \centering
    \includegraphics[width=\linewidth]{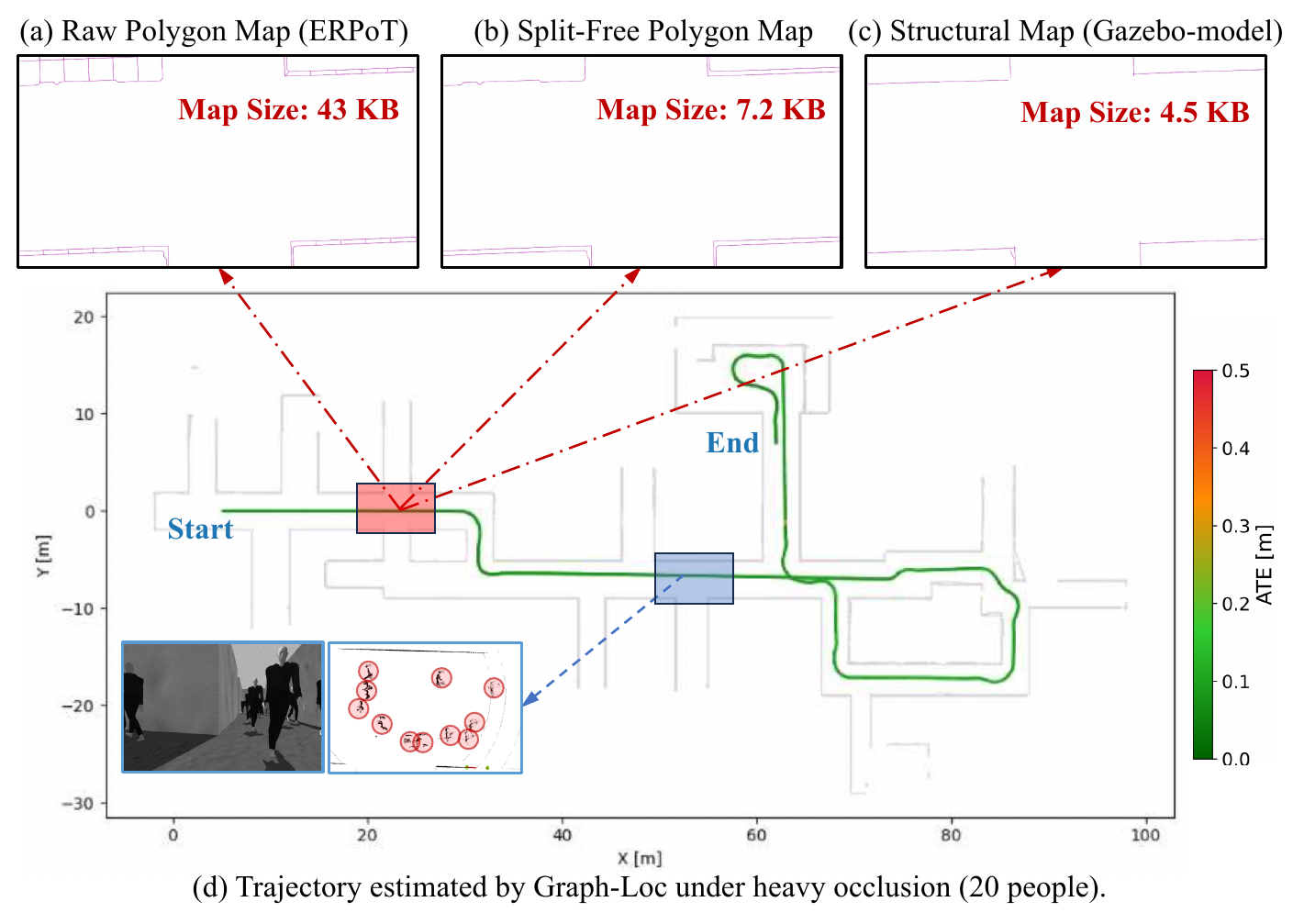}
    \caption{Qualitative results on CMU-EXPLORATION under the heavy-occlusion setting.
    (a) Raw polygon map used by ERPoT, where long structural outlines are retained in the original occupancy-derived representation.
    (b) Split-free polygon map used by Graph-Loc, which preserves compact outlines without additional contour segmentation.
    (c) Structural map distilled from the Gazebo model, providing an alternative compact layout prior.
    (d) Trajectory estimated by Graph-Loc using the split-free polygon prior under heavy occlusion with 20 simulated pedestrians.}
    \label{Fig_7}
\end{figure}

\textbf{Priors and baselines.}
We report results under two compact prior sources: (i) a structural layout prior distilled from the Gazebo model, and (ii) polygon outlines vectorized from occupancy/grid-style maps.
For the polygon prior, Graph-Loc keeps the outlines unsplit to preserve compactness.
In this weakly constrained setting, the original ERPoT polygon pipeline is not directly runnable.
To include ERPoT as a reference, we additionally apply a length-based splitting rule beyond its default segmentation to break long polygons into shorter primitives.
With this modification, ERPoT becomes trackable only in the static setting, while requiring a substantially larger polygon map (about $6\times$ compared to the unsplit polygon prior); under pedestrian-induced occlusions, it remains unstable and does not provide reliable tracking.

\textbf{Results and analysis.}
Table~\ref{tab:sim_results} reports quantitative results and Figure~\ref{Fig_7} visualizes representative trajectories and map constructions.
With the Gazebo-derived structural prior, the resulting map provides clean and stable constraints, yielding reliable tracking in the static and mild-occlusion settings.
With the unsplit polygon-outline prior, Graph-Loc maintains stable trajectory alignment as occlusion increases, indicating that compact occupancy-derived priors can remain effective without contour splitting when association is solved globally and pose updates are stabilized by degeneracy-aware refinement.

\rev{Overall, this controlled study complements DOALS by explicitly varying pedestrian-induced occlusion density while keeping the route, prior map, and structural layout fixed.
It shows that Graph-Loc can maintain reliable compact-prior tracking under controlled dynamic occlusions in a weakly constrained environment, and further demonstrates that the proposed framework supports heterogeneous compact priors, including occupancy-derived outlines and layout-based priors.}

\begin{figure}
    \centering
    \includegraphics[width=\linewidth]{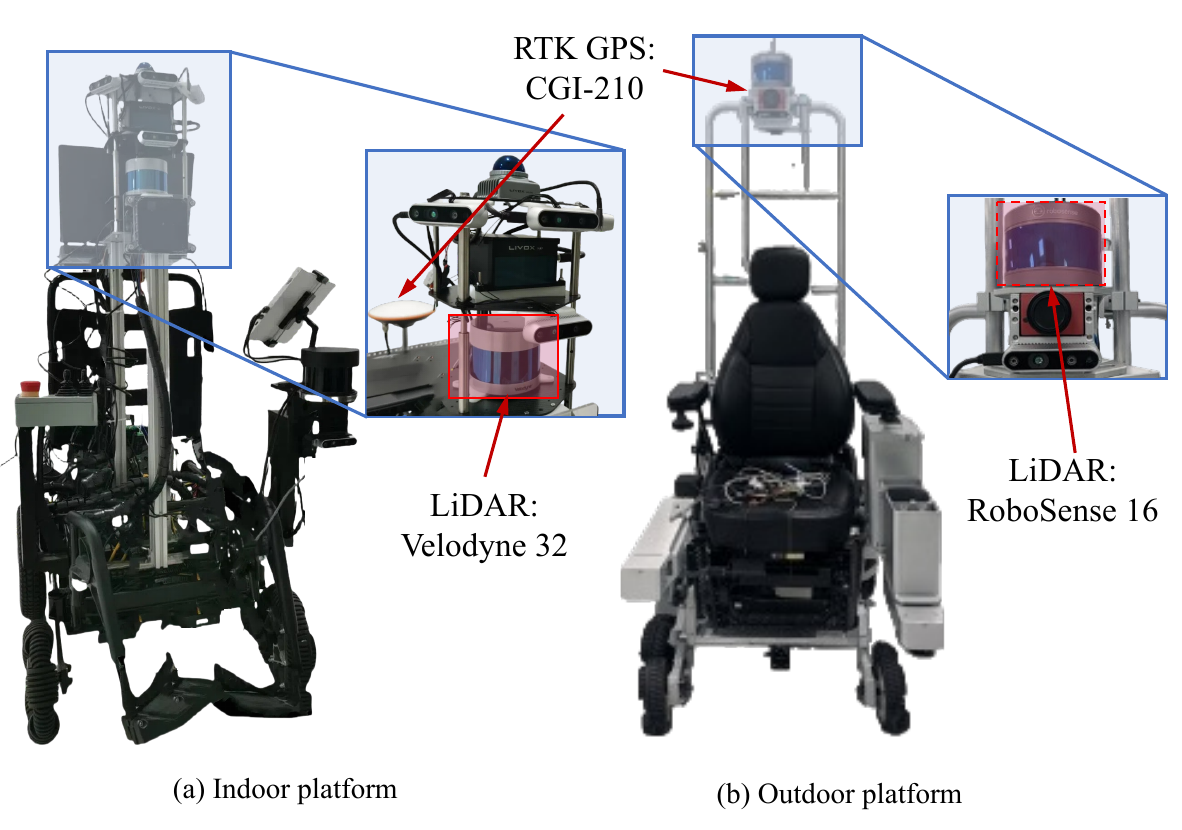}
    \caption{Our data-collection platforms.}
    \label{Fig_8}
\end{figure}

\subsection{\rev{Self-Collected Mobile Robot Deployments}}
In this section, we conduct real-world experiments to validate deployable pose tracking with compact priors under two practical factors: (i) weak observability with sustained short-range occlusions, and (ii) long-range tracking under gradual scene changes.
Figure~\ref{Fig_8} shows the two data-collection platforms.
The first setup targets ground navigation in a corridor-like route using a Velodyne 32-beam LiDAR, where observations are often dominated by repeated straight structures and frequently interrupted by nearby occluders.
The second setup targets a parking-lot traversal using a RoboSense 16-beam LiDAR, where the map prior can be imperfect due to parked vehicles, temporary obstacles, and local layout variations across days.
For evaluation, outdoor ground truth is provided by RTK-GPS, while indoor reference trajectories are produced by an offline LIVO pipeline with dynamic filtering and loop-closure optimization.

\begin{figure}[t]
    \centering
    \includegraphics[width=\linewidth]{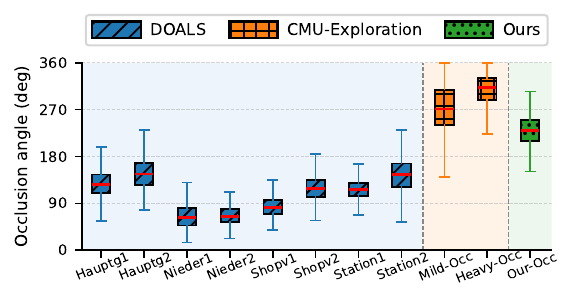}
    \caption{Horizontal occlusion-angle distributions across dynamic-obstruction evaluation sequences.
    DOALS represents natural pass-by pedestrian occlusions, whereas CMU-EXPLORATION and the indoor deployment include pedestrians moving with or around the robot, producing more persistent azimuth-level blockage.
    The distributions characterize planar blockage patterns across complementary dynamic-obstruction regimes, indicating that occlusion difficulty is related to blockage pattern rather than pedestrian count alone.}
    \label{Fig_9}
\end{figure}

\begin{table*}[t]
\centering
\caption{Pose-tracking results on real degeneracy and occlusion.}
\label{tab:real_results}
\begin{tabular}{c l c *{6}{c}}
\toprule
\multirow{2}{*}{Map Type} &
\multirow{2}{*}{Method} &
\multirow{2}{*}{\begin{tabular}[c]{@{}c@{}}Prior Map\\ Size (KB) $\downarrow$\end{tabular}} &
\multicolumn{3}{c}{Static (ATE [cm])} &
\multicolumn{3}{c}{Occlusion (ATE [cm])} \\
\cmidrule(l{2pt}r{2pt}){4-6}
\cmidrule(l{2pt}r{2pt}){7-9}
& & &
Max $\downarrow$ & Mean $\downarrow$ & RMSE $\downarrow$ &
Max $\downarrow$ & Mean $\downarrow$ & RMSE $\downarrow$ \\
\hline
\multirow{2}{*}{\begin{tabular}[c]{@{}c@{}}Point\\Cloud\end{tabular}}
& HDL\_LOC    
& 2682.88      
& 47.64 & 22.40 & 25.06
& 50.20 & 22.83 & 25.33 \\

& ALOAM\_MCL   
& 5365.76 
& 43.68 & 19.54 & 21.82
& \underline{50.07} & 20.40 & 22.50 \\
\noalign{\vskip 2pt}\cdashline{1-9}\noalign{\vskip 2pt}
\multirow{2}{*}{CAD}
& PS\_LOC        
& 24.8     
& 48.15 & 18.27 & 20.74
& 51.23 & 22.35 & 23.21  \\

& \cellcolor{gray!15}Ours        
& \cellcolor{gray!15}24.8 
& \cellcolor{gray!15}\textbf{30.91} & \cellcolor{gray!15}\textbf{11.47} & \cellcolor{gray!15}\textbf{13.76}
& \cellcolor{gray!15}58.98 & \cellcolor{gray!15}\underline{14.04} & \cellcolor{gray!15}\underline{17.80} \\
\noalign{\vskip 2pt}\cdashline{1-9}\noalign{\vskip 2pt}
\multirow{2}{*}{\begin{tabular}[c]{@{}c@{}}Polygon\\Map\end{tabular}}
& ERPoT        
& 184     
& 61.26 & 15.20 & 18.08
& \NA   & \NA   & \NA  \\

& \cellcolor{gray!15}Ours        
& \cellcolor{gray!15}154 
& \cellcolor{gray!15}\underline{35.21} & \cellcolor{gray!15}\underline{12.80} & \cellcolor{gray!15}\underline{15.29}
& \cellcolor{gray!15}\textbf{30.28} & \cellcolor{gray!15}\textbf{13.80} & \cellcolor{gray!15}\textbf{15.00} \\
\bottomrule
\end{tabular}
\end{table*}

\begin{figure}[t]
    \centering
    \includegraphics[width=\linewidth]{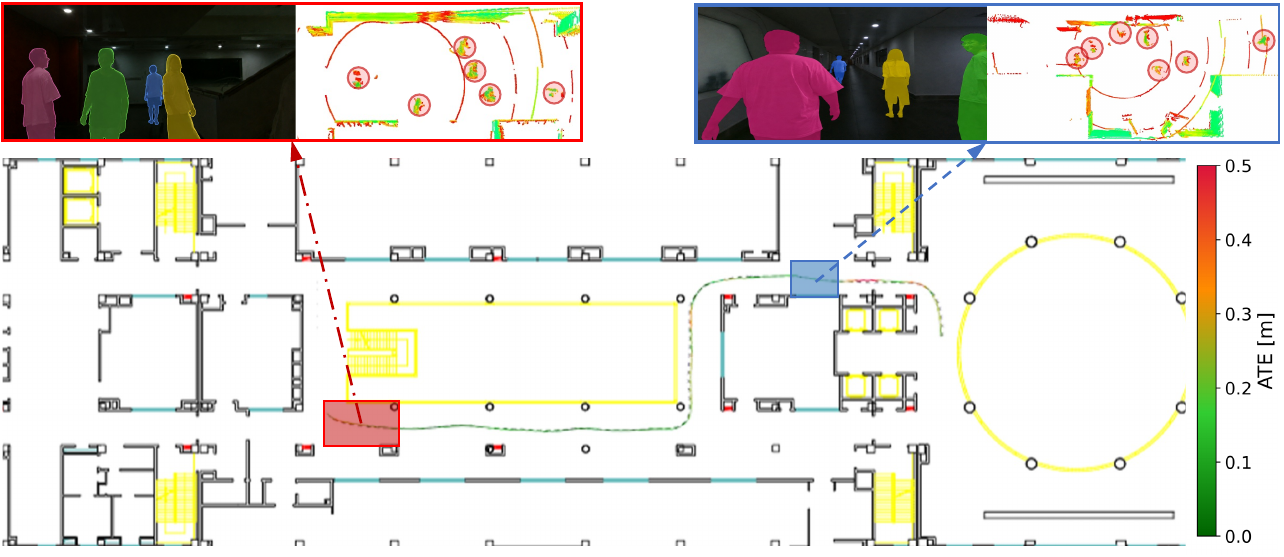}
    \caption{Trajectory estimated by Graph-Loc using the CAD prior under sustained close-range pedestrian occlusion.}
    \label{Fig_10}
\end{figure}

\subsubsection{\rev{Degeneracy and Occlusion}}
\rev{Figure~\ref{Fig_9} compares the occlusion-angle distributions across DOALS, CMU-EXPLORATION, and our indoor occlusion sequence.
The occlusion angle is measured as the horizontal angular span of LiDAR rays affected by dynamic or transient objects after projecting the scan and occluder regions onto the ground plane.
This planar metric characterizes azimuth-level blockage patterns rather than full 3D LiDAR visibility, and is used as a diagnostic descriptor of dynamic obstruction instead of a direct predictor of localization failure.
The comparison shows that pedestrian count alone does not characterize occlusion difficulty: DOALS mainly contains natural pass-by pedestrian occlusions where fixed structural regions remain repeatedly observable, whereas CMU-EXPLORATION and our indoor occlusion sequence involve pedestrians moving with or around the robot, producing more persistent field-of-view blockage over consecutive scans.
Thus, the indoor occlusion sequence complements the public and simulated dynamic-obstruction settings by evaluating practical close-range occlusion in a real mobile-robot scenario.}

Two indoor deployment sequences are collected in the same site: a static sequence and an occlusion sequence with sustained close-range pedestrian interference.
The occlusion sequence is implemented by five pedestrians moving with or around the platform, creating persistent field-of-view blockage over consecutive scans.
The prior map is built from an independent mapping session recorded on a different day.
As reported in Table~\ref{tab:real_results}, the proposed method achieves the best overall accuracy across both sequences while running online in real time.
With the CAD prior in the static sequence, the clean layout cues enable reliable scan-to-map alignment, and the proposed global graph matching improves accuracy over the CAD-based baseline.
With pedestrian interference, a non-negligible portion of measurements becomes missing or corrupted; nevertheless, the proposed method remains stable and consistently outperforms both point-cloud-prior and polygon-prior baselines.
Figure~\ref{Fig_10} visualizes the trajectory on the occlusion sequence with the CAD prior, illustrating accurate real-time tracking under sustained close-range pedestrian occlusion.

\begin{table}[t]
\centering
\caption{Pose-tracking results under gradual environment changes.}
\label{tab:outdoor_results}
\begin{tabular}{l c c c c}
\toprule
\multirow{2}{*}{Method} &
\multirow{2}{*}{\begin{tabular}[c]{@{}c@{}}Prior Map\\ Size (MB) $\downarrow$\end{tabular}} &
\multicolumn{3}{c}{ATE [cm]} \\
\cmidrule(l{2pt}r{2pt}){3-5}
& &
Max $\downarrow$ & Mean $\downarrow$ & RMSE $\downarrow$ \\
\midrule

HDL\_LOC     & 3.0 & \underline{65.49} & \underline{11.10} & \underline{12.83} \\
ALOAM\_MCL   & 3.0 & \NA & \NA & \NA \\

ERPoT        & 0.15 & 110.6 & 27.79 & 33.69 \\
Ours        & 0.13 & \textbf{46.63} & \textbf{5.18} & \textbf{7.42} \\

\bottomrule
\end{tabular}
\end{table}

\begin{figure}[t]
    \centering
    \includegraphics[width=\linewidth]{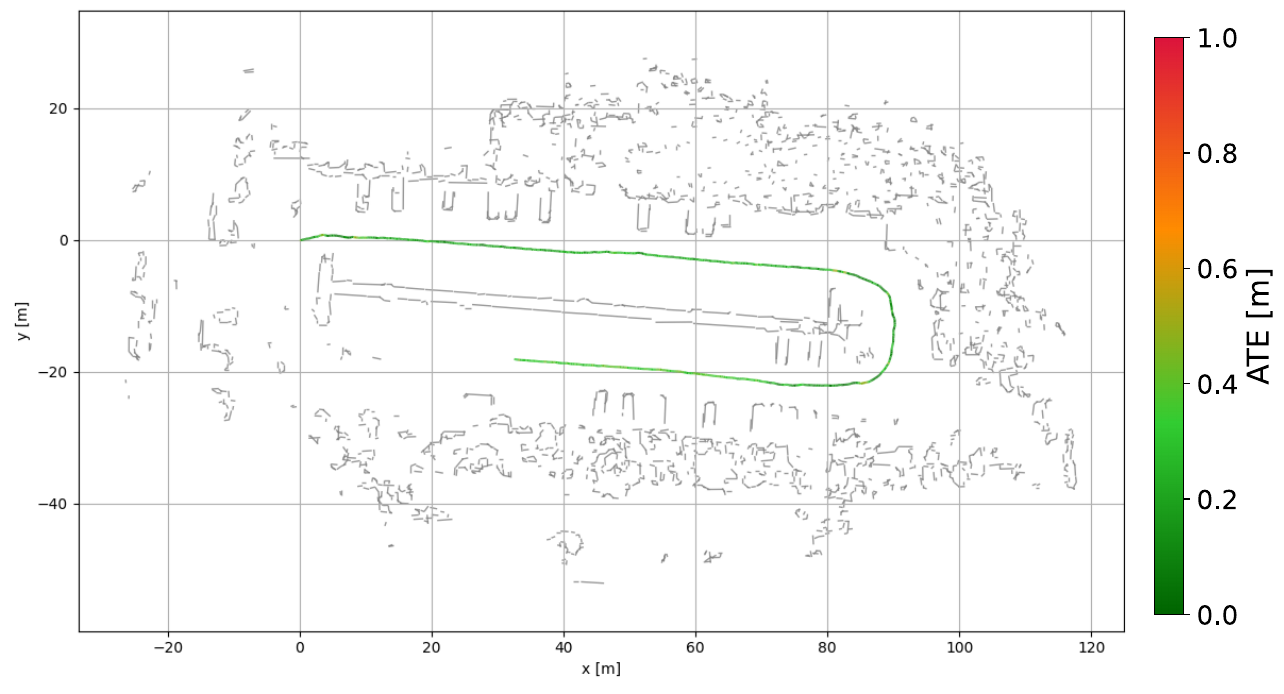}
    \caption{Trajectory estimated by Graph-Loc using the compact polygon-outline prior under gradual parking-lot environment changes.}
    \label{Fig_11}
\end{figure}

\begin{figure*}[t]
    \centering
    \includegraphics[width=\linewidth]{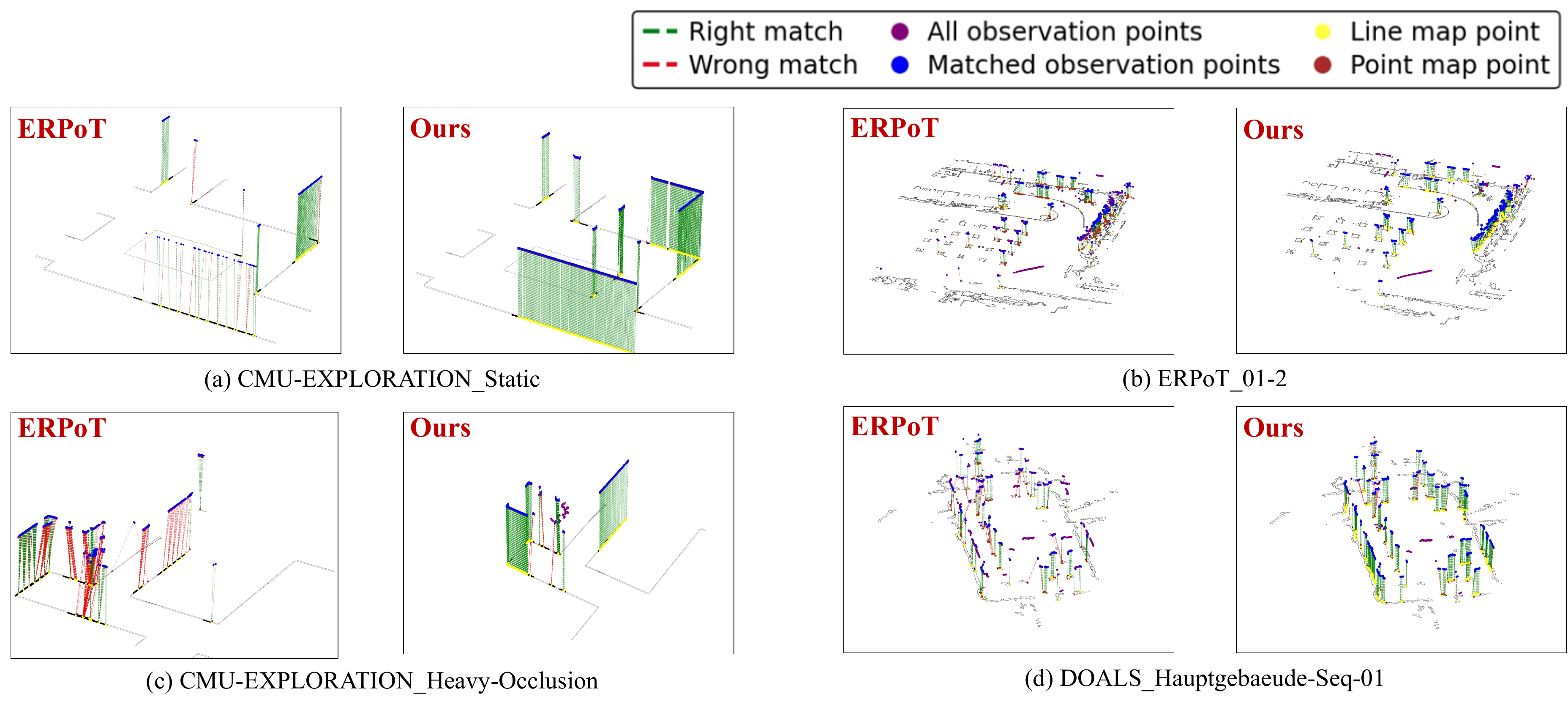}
    \caption{\rev{Qualitative comparison of feature association between ERPoT and Graph-Loc.
    Subfigures (a) and (b) are static scenes, while (c) and (d) are dynamic scenes; the CMU-EXPLORATION cases in (a) and (c) additionally represent structurally degenerate settings.
    Graph-Loc yields more reliable point-line correspondences, especially for line matching, under both structural degeneracy and dynamic interference.}}
    \label{Fig_12}
\end{figure*}

\subsubsection{Tracking under environment changes}
We further evaluate long-range tracking on a parking-lot sequence with noticeable environment changes across time.
Table~\ref{tab:outdoor_results} and Figure~\ref{Fig_11} report results using point-cloud priors and polygon-outline priors.
Graph-Loc achieves the best accuracy among all compared approaches, obtaining the lowest ATE while using a compact polygon-outline prior.
In particular, compared with the split-based polygon pipeline ERPoT, Graph-Loc substantially reduces the trajectory error and remains stable under partial visibility and prior mismatch caused by parked vehicles and temporary obstacles.
Compared with the point-cloud baseline HDL\_LOC, Graph-Loc also yields lower ATE across all metrics, while requiring a much smaller prior map.
Overall, the outdoor results indicate that compact polygon priors can support accurate and stable long-range localization under gradual scene changes, without contour splitting or map inflation.

\subsection{Ablation Study}
\label{sec:ablation}

Table~\ref{tab:ablation} reports an ablation study on KITTI, ERPoT, and CMU-EXPLORATION.
We ablate three key components: dynamic filtering (D.F.), unbalanced optimal transport matching (O.T.), and degeneracy-aware delayed optimization (D.T.).

\begin{table}[t]
\centering
\caption{Ablation study of different components on KITTI, ERPoT, and CMU-EXPLORATION (RMSE [cm]). D.F. denotes dynamic feature filtering, O.T. denotes correspondence estimation via unbalanced optimal-transport graph matching, and D.T. denotes degeneracy-aware delayed optimization.}
\label{tab:ablation}
\resizebox{\linewidth}{!}{
\begin{tabular}{l|cccc}
\hline
Sequence    & w.o. D.F. & w.o. O.T. & w.o. D.T. & Full \\ \hline
KITTI00-1   & 11.2   & 21.4   & 13.3   & 10.9 \\
KITTI00-2   & 12.1   & 26.4   & 14.1   & 11.3 \\
KITTI00-3   & 10.7   & 23.4   & 14.8   & 9.6  \\
KITTI02-1   & 17.2   & 17.4   & 16.3   & 15.5 \\
KITTI02-2   & 17.6   & 16.1   & 20.5   & 15.9 \\
KITTI05-1   & 12.3   & 13.4   & 11.6   & 10.3 \\
KITTI05-2   & 14.1   & 16.1   & 10.9   & 10.4 \\
KITTI06-1   & 10.2   & \NA    & 8.0    & 7.3  \\
KITTI08-1   & 22.1   & 25.7   & 21.6   & 21.3 \\ \hdashline
ERPoT-01-1  & 6.8    & 8.9    & 8.0    & 6.6  \\
ERPoT-01-2  & 6.7    & 8.6    & 5.6    & 5.5  \\
ERPoT-02-1  & 13.2   & 13.8   & 17.3   & 10.6 \\ \hdashline
CMU-Static      & 8.8    & 12.0   & 25.7   & 8.1  \\
CMU-Mild      & 11.3   & \NA    & 28.0   & 9.9  \\
CMU-Heavy     & 17.3   & \NA    & 30.9   & 10.6 \\ \hline
\end{tabular}
}
\end{table}

Across datasets, the \textbf{unbalanced optimal transport} module is the most critical.
Removing it causes the largest and most consistent degradation on both KITTI and ERPoT, even though these benchmarks are static or near-static.
In the ablated setting without unbalanced optimal transport, correspondences are computed using the same nearest-neighbor association scheme as ERPoT~\cite{gao2025erpot}, i.e., each feature selects its closest candidate under the current pose estimate and the pose is refined by iterative re-alignment.
The clear performance gap indicates that local nearest-neighbor decisions are particularly brittle in repetitive or weakly distinctive structures, whereas the global transport plan provides substantially more reliable association signals for pose tracking.

The \textbf{degeneracy-aware delayed optimization} becomes especially important in structurally degenerate segments.
While robust matching provides reliable correspondences, the pose update can still be ill-conditioned when the visible structure offers weak constraints, such as long corridors or dominant parallel boundaries.
This effect is most evident in CMU-EXPLORATION: disabling the delayed optimization leads to a pronounced performance drop, and the gap grows as occlusion increases from CMU-Static to CMU-Heavy.
By damping updates along weakly observable directions and postponing aggressive corrections until sufficient constraints accumulate, the delayed optimization stabilizes tracking under low-observability conditions.

The \textbf{dynamic filtering} module mainly improves robustness under dynamics.
Since KITTI, ERPoT, and CMU-Static are static or quasi-static, removing dynamic filtering yields a comparatively smaller change there.
Under dynamic occlusions, however, transient returns and fragmented observations increase the burden on association and pose refinement.
By suppressing unreliable dynamic features before association, dynamic filtering preserves cleaner structural cues and reduces the burden on subsequent graph association and optimization.

\textbf{Qualitative association analysis.}
\rev{Figure~\ref{Fig_12} further compares feature association between ERPoT and Graph-Loc in representative static and dynamic scenes.
Subfigures (a) and (b) are static scenes, while (c) and (d) are dynamic scenes; the CMU-EXPLORATION cases in (a) and (c) additionally represent structurally degenerate settings.
For visualization, a correspondence is marked as correct if the observation feature transformed by the ground-truth pose is geometrically consistent with the matched map feature within a preset distance threshold.
We use 0.3~m for indoor or compact scenes and 1.0~m for outdoor large-scale scenes; these thresholds are used only to evaluate and color the displayed correspondences.
The comparison shows that Graph-Loc produces more reliable point-line correspondences, especially more stable line matching, under both structural degeneracy and dynamic interference.
This qualitative behavior is consistent with the ablation results: dynamic filtering suppresses unreliable dynamic observations before association, while unbalanced optimal transport provides globally consistent matching instead of local nearest-neighbor decisions.}

Overall, the ablation supports a clear division of roles: optimal transport matching provides the core robust association capability, dynamic filtering suppresses transient returns under dynamics, and delayed optimization stabilizes pose refinement under low-observability conditions.

\subsection{Runtime}
\label{sec:runtime}

As shown in Table~\ref{Tab_7}, we report runtime on the CMU-EXPLORATION benchmark by running 10 independent trials and measuring the per-scan processing time for each method.
All methods are run on a PC equipped with an Intel i5-14600KF CPU and an NVIDIA RTX 5060 Ti GPU.
Overall, our GPU implementation achieves the lowest mean latency and a substantially tighter worst-case bound than ERPoT, while the CPU implementation remains competitive in mean time but exhibits a larger variance.
A key reason is that the optimal transport solver is particularly amenable to GPU acceleration.
Its Sinkhorn-style iterations are dominated by element-wise operations and matrix-vector/matrix-matrix primitives, which map efficiently to GPU kernels.
In contrast, ERPoT relies on thresholded association followed by heuristic iterative refinement and pruning.
By solving correspondences globally and typically converging to a stable transport plan in a small, fixed number of iterations, the proposed matching reaches stable associations faster in practice, leading to the runtime advantage.

\begin{table}[t]
\centering
\caption{Runtime comparison.}
\renewcommand{\arraystretch}{1.2}
\begin{tabular}{lcccc}
\toprule
Method            & Max [ms] & Mean [ms] & Min [ms] & SD [ms] \\
\midrule
ERPoT             & 162.00 & 100.17 & 51.00 & 21.60 \\
Ours (CPU)  & 394.00 &  63.91 &  1.00 & 45.50 \\
Ours (GPU)  &  62.00 &  24.56 &  6.00 &  6.86 \\
\bottomrule
\end{tabular}
\label{Tab_7}
\end{table}

\section{Conclusion}
We presented Graph-Loc, a graph-based LiDAR pose-tracking framework that localizes against compact structural priors represented as lightweight point-line graphs.
While our primary focus is on split-free polygon outlines vectorized from occupancy/grid-style maps, Graph-Loc also supports heterogeneous compact priors commonly available in practice, including CAD/model/floor-plan layouts in indoor scenarios.
Scan-to-map association is solved globally via unbalanced optimal transport with graph-context regularization, reducing the reliance on offline contour segmentation that inflates the map to improve matchability.
The unbalanced formulation relaxes strict mass conservation, making matching tolerant to missing, spurious, and fragmented observations under occlusion and partial visibility.
To stabilize pose refinement in low-observability segments (e.g., corridor-like layouts dominated by parallel structures), we further introduced a degeneracy-aware delayed optimization strategy that detects weakly constrained motion directions, accumulates reliable constraints, and releases full updates once observability recovers.
Experiments on KITTI, ERPoT, and MulRan show that split-free compact structural priors can achieve competitive tracking accuracy while consistently reducing prior size and supporting large-scale long-term repeated-route localization.
\rev{Controlled simulation analyzes robustness trends under systematically varied occlusion and low-observability conditions, while DOALS and self-collected field deployments validate practical robustness under natural dynamic disturbances, sustained occlusion, and gradual scene changes.}

\bibliographystyle{IEEEtran}
\bibliography{main}

\vfill

\end{document}